\documentclass[10pt,twocolumn,letterpaper]{article}

\usepackage[pagenumbers]{cvpr} 

\usepackage{multirow}
\usepackage{marvosym}

\definecolor{cvprblue}{rgb}{0.21,0.49,0.74}
\usepackage[pagebackref,breaklinks,colorlinks,allcolors=cvprblue]{hyperref}

\def\paperID{8707} 
\def\confName{CVPR}
\def\confYear{2025}

\title{FATE: Full-head Gaussian Avatar with Textural Editing \\ from Monocular Video}

\author{
Jiawei Zhang\textsuperscript{\rm 1}
\quad
Zijian Wu\textsuperscript{\rm 1}
\quad
Zhiyang Liang\textsuperscript{\rm 1}
\quad
Yicheng Gong\textsuperscript{\rm 1}
\\
Dongfang Hu\textsuperscript{\rm 2}
\quad
Yao Yao\textsuperscript{\rm 1}
\quad
Xun Cao\textsuperscript{\rm 1}
\quad
Hao Zhu\textsuperscript{\rm 1,\Letter }
\vspace{5pt}\\
\textsuperscript{\rm 1}Nanjing University
\qquad
\textsuperscript{\rm 2}OPPO
}

\definecolor{myPurple}{rgb}{0.4, .0, .8}
\definecolor{myGreen}{rgb}{0, .8, .3}
\definecolor{myRed}{rgb}{0.8, .2, .2}
\definecolor{myOrange}{rgb}{0.8, 0.45, 0.0}
\definecolor{myBlue}{rgb}{.0, .0, 1.0}

\usepackage{booktabs}
\usepackage{colortbl}
\newcommand{\cb}{\cellcolor{blue!50}}
\newcommand{\clb}{\cellcolor{blue!20}}
\usepackage{threeparttable}

\begin{document}

\twocolumn[{%
\renewcommand\twocolumn[1][]{#1}%
\maketitle
\begin{center}
    \centering
    \captionsetup{type=figure}
    \vspace{-0.3in}
     \includegraphics[width=\textwidth]{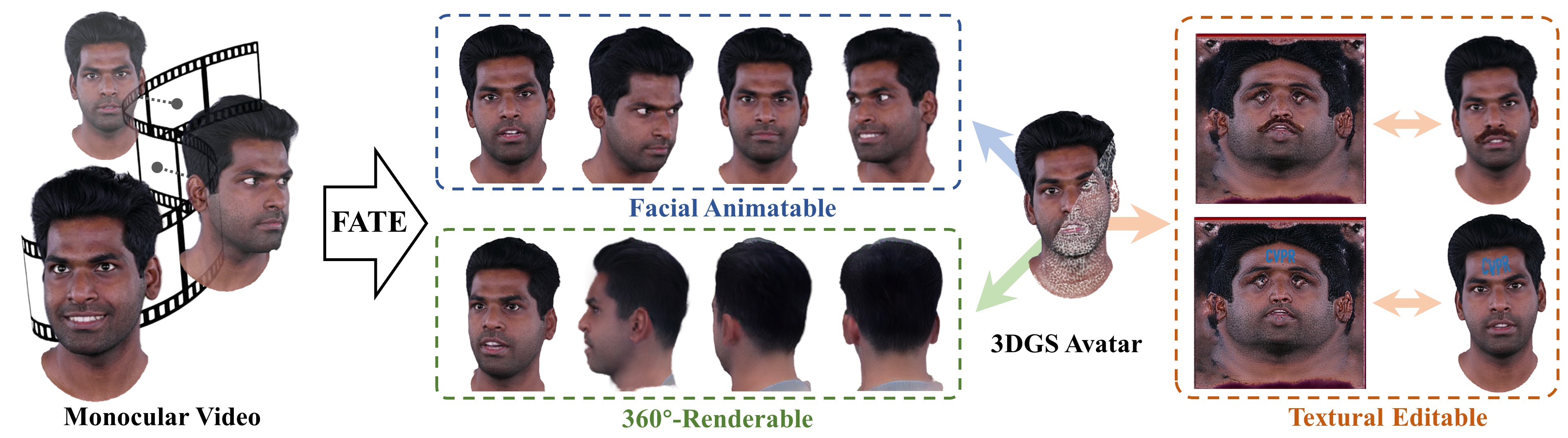}
     \vspace{-0.2in}
      \caption{
      From a monocular portrait video input, we propose FATE to reconstruct an animatable 3D head avatar, which enables Gaussian texture editing and allows for 360$^\circ$ full-head synthesis.
      }
      \vspace{0.2in}
      \label{fig:title}
\end{center}%
}]

\maketitle

\begin{abstract}
Reconstructing high-fidelity, animatable 3D head avatars from effortlessly captured monocular videos is a pivotal yet formidable challenge.  
Although significant progress has been made in rendering performance and manipulation capabilities, notable challenges remain, including incomplete reconstruction and inefficient Gaussian representation.
To address these challenges, we introduce FATE — a novel method for reconstructing an editable full-head avatar from a single monocular video.  FATE integrates a sampling-based densification strategy to ensure optimal positional distribution of points, improving rendering efficiency.  A neural baking technique is introduced to convert discrete Gaussian representations into continuous attribute maps, facilitating intuitive appearance editing.  Furthermore, we propose a universal completion framework to recover non-frontal appearance, culminating in a 360$^\circ$-renderable 3D head avatar.
FATE outperforms previous approaches in both qualitative and quantitative evaluations, achieving state-of-the-art performance.  To the best of our knowledge, FATE is the first animatable and 360$^\circ$ full-head monocular reconstruction method for a 3D head avatar. Project page and code are available at this \href{https://zjwfufu.github.io/FATE-page/}{link}.
\end{abstract}    
\section{Introduction}
\label{sec:intro}

Reconstructing photo-realistic and animatable 3D head avatars is a consistent objective in computer vision, given its extensive applications in film production, AR/VR, meta-verse, and computer games.  To produce high-fidelity head avatars with precision, classic solutions commonly rely on light field acquisition systems~\cite{debevec2012light, guo2019relightables, yang2023towards} alongside the design of an artist team. These approaches require huge costs and unvoidable manual design, which can hardly be applied to consumer-level scenarios.  In recent years, significant research efforts have been devoted to a more practical approach: reconstructing 3D head avatars from an easily captured monocular video.

Early research on the monocular reconstruction of 3D head avatars converges to a widely adopted framework.  Firstly, parametric head estimation algorithms~\cite{MICA:ECCV2022, DECASiggraph2021, EMOCACVPR:2021} are leveraged to estimate a head's pose and rough shape for each frame.  Subsequently, multiple video frames are harnessed to refine the head's appearance across various poses and expressions, culminating in an expression-drivable 3D head avatar.  The advent of the 3D Gaussian Splatting (3DGS)~\cite{kerbl3Dgaussians} model, renowned for its rendering efficiency and ease of manipulation, has been widely adopted as the preferred head representation in recent methods~\cite{qian2023gaussianavatars, saito2024rgca, xiang2024flashavatar, xu2023gaussianheadavatar, SplattingAvatar:CVPR2024}. Despite significant performance advancements, monocular 3D head avatar reconstruction still confronts several unresolved challenges.

The first issue is incompleteness in head modeling.  Previous approaches predominantly focus on modeling the frontal human face and fail to recover the rear head.  This limitation is rooted in the reliance on parametric face estimation methods.  Specifically, due to the lack of facial features, both landmark-based and landmark-free parametric head estimation methods fail for the rear head.  Thus, video frames of the rear head can not be used in the following optimization process.  Practically, most portrait videos focus on informative frontal imagery, with the less informative rear views being scarcely captured.  Recovering 360-$^\circ$ full 3D head from frontal videos remains an unsolved challenge.

The second issue pertains to the inefficiency and discreteness of the 3DGS representations. We observed that the densification mechanism inherent to the original 3DGS model is ill-suited for monocular reconstruction tasks, as it produces a plethora of redundant attributed points in the training stage. These redundant points compromise rendering quality and increase model complexity. Moreover, due to the discrete nature of the 3D Gaussian representation, the 3DGS-represented head can not be directly edited in the UV texture space, just like polygon mesh models. Previous editable methods~\cite{song2024texttoon, Gao2024PortraitGen, bao2024geneavatar} rely on extensive optimization with pre-trained diffusion models~\cite{zhang2024realcompo,zhang2024itercompo}, such as InstructPix2Pix~\cite{brooks2022instructpix2pix}, which is both time-consuming and uncontrollable. Although some prior methods~\cite{xiang2024flashavatar, zhao2024psavatar, SplattingAvatar:CVPR2024, GSM, kirschstein2024gghead} also structure Gaussian points into the UV space, our experiments reveal that their reconstructed textures are discontinuous in the UV domain.

To solve these challenges, we introduce FATE, a novel method to reconstruct an editable and full-head avatar from a monocular video. To tackle the problem of model inefficiency, we propose a sampling-based densification approach that achieves a more optimal position distribution than previous methods. Furthermore, we devise a novel technique for parameterizing trained Gaussian points in UV space into multiple attribute maps, thereby enabling the editing of Gaussians with the same ease as mesh textures. To resolve the challenge of reconstructing a fully 360$^\circ$ renderable head, we develop a universal completion framework that extracts appearance-customized priors from SphereHead~\cite{li2024spherehead}, a pre-trained generative model. This framework is not only compatible with our FATE method, but can also be seamlessly integrated into other head reconstruction methods~\cite{xiang2024flashavatar, chen2024monogaussianavatar, qian2023gaussianavatars, SplattingAvatar:CVPR2024}. The FATE model outperforms state-of-the-art methods in qualitative and quantitative evaluations. To the best of our knowledge, FATE is the first animatable and 360$^\circ$ full-head monocular reconstruction method for a 3D head avatar.

Our contributions can be summarized as:

\begin{itemize}
\item We propose a monocular video reconstruction method incorporating \textit{sampling-based densification}. Comprehensive experiments demonstrate that our method attains state-of-the-art qualitative and quantitative results.

\item Neural baking is introduced to transform discrete Gaussian representations onto continuous attribute maps in the UV space. This enables appearance editing with the same ease and efficacy as mesh textures.

\item We propose the first and universal completion framework that improves the reconstruction of non-frontal viewpoints by acquiring priors from a pre-trained generative model, leading to a fully 360$^\circ$-renderable 3D head avatar from a monocular video.
\end{itemize}
\section{Related Work}
\label{sec:related}


\begin{figure*}[htbp!]
    \centering
    \includegraphics[width=\textwidth]{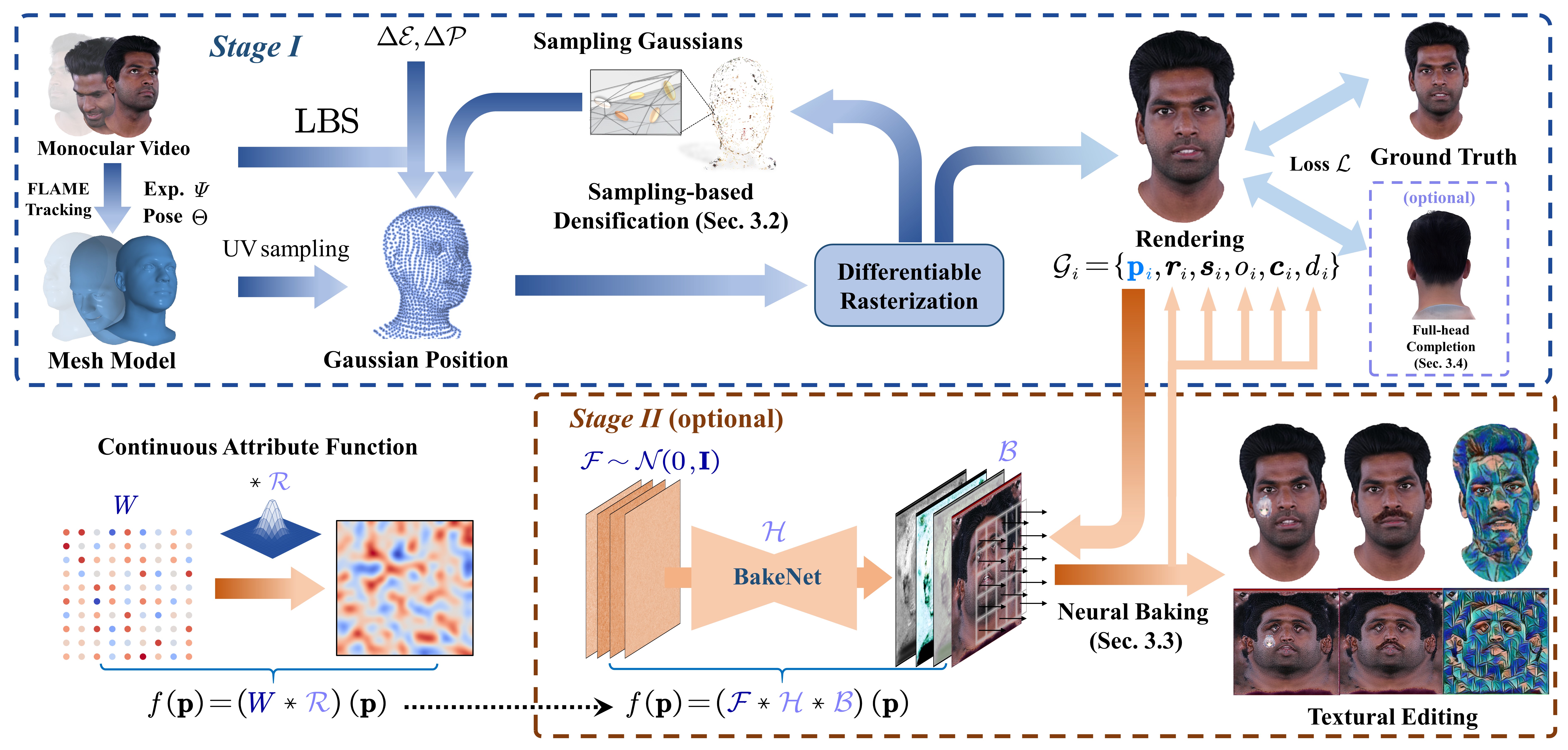}
    \caption{\textbf{Pipeline}. In \textit{Stage I}, we perform sampling-based densification in Sec.~\ref{sec: densify} in the UV space and train a Gaussian head avatar using the preprocessed monocular video dataset.  The obtained head avatar can optionally use full-head completion in Sec~\ref{sec: aug} to recover non-frontal regions. In \textit{Stage II}, given the learned head avatar, we construct a continuous function $f(\mathbf{p})$ in the UV space using U-Net $\mathcal{H}$ and bilinear kernel $\mathcal{B}$, baking the Gaussian attributes into several maps as described in Sec \ref{sec: baking}. 
    }
    \label{fig: pipeline}
    \vspace{-18px}
\end{figure*}

\subsection{Monocular Head Avatar Reconstruction}

Recovering a 3D head avatar from a monocular video is a very ill-posed problem, considering unconstrained head pose and deformation. To regularize the problem, most approaches resort to 3D Morphable Models (3DMM)~\cite{BFM, FaceWarehouse, FLAME, zhu2023facescape, he2024emotalk3d} as geometric knowledge, by which expression and pose parameters for each video frame are estimated using either a learning-based decoder~\cite{DECASiggraph2021, EMOCACVPR:2021, deng2019accurate} or an optimization-based face tracker~\cite{MICA:ECCV2022}. These coefficients serve as conditions or driving signals to facilitate head reconstruction.

The emergence of NeRF has sparked a growing interest in the implicit modeling of head avatars through ray-casting techniques.  By conditioning on expression and pose, several works~\cite{zheng2022imavatar, Gafni_2021_CVPR, wang2025uarnvcunifiedautoregressiveframework, zhuang2023neai, bakedavatar, NHA, INSTA} learn a deformation field for animatable 3D head avatar. 
NerFACE \cite{Gafni_2021_CVPR} utilizes FLAME coefficients as a condition and feeds them into MLP to synthesize dynamic avatars. IMavatar \cite{zheng2022imavatar} proposes to learn head avatars with implicit geometry and texture model, providing novel analytical gradient formulation that enables end-to-end training from videos. 
BakedAvatar \cite{bakedavatar} utilizes deformable multi-layer meshes in head avatar reconstruction to improve rendering. Though significantly enhanced in rendering quality, the NeRF-based method requires pixel-by-pixel ray casting and queries from a multilayer perceptron (MLP), considerably limiting its training and inference efficiency. Latter works~\cite{xu2023avatarmav, Gao2022nerfblendshape, gao2024mani, song2024tri, INSTA} have employed voxel hashing~\cite{mueller2022instant} or tensor decomposition~\cite{Chen2022ECCV} to accelerate this process, achieving varying degrees of success.

Recently, 3D Gaussian Splatting (3DGS) has garnered significant attention. 3DGS represents scenes using numerous anisotropic Gaussian splats, each characterized by geometry and appearance attributes. This explicit modeling method is fast and highly controllable, leading to multiple real-time and high-fidelity avatar reconstruction methods. One track is to use high-cost multi-view datasets and involves complex designs to achieve ultra-rendering quality.  
RGCA~\cite{saito2024rgca} uses a conditional variational autoencoder to learn Gaussian attributes and radiance transfer. 
Gaussian Head Avatar~\cite{xu2023gaussianheadavatar} first obtains SDF-based geometry from multi-view videos and then achieves high-resolution rendering under deformed MLPs and a super-resolution network. 
GaussianAvatars~\cite{qian2023gaussianavatars} applies a binding mechanism to attach Gaussians to the mesh faces. 

As for monocular video, FlashAvatar~\cite{xiang2024flashavatar} obtains a high-fidelity head avatar by uniform UV sampling. PSAvatar~\cite{zhao2024psavatar} spreads dense Gaussian points on and off the mesh to facilitate detailed capture. 
SplattingAvatar~\cite{SplattingAvatar:CVPR2024} makes Gaussians walk along triangles to enhance the representation.
GaussianBlendshapes~\cite{ma2024gaussianblendshapes} proposes to build Gaussian attribute basis referring to blendshapes. 
MonoGaussianAvatar~\cite{chen2024monogaussianavatar} leverages MLPs to predict Gaussian attributes and designs a scale and sampling scheduler to enable progressive training. While these methods have achieved commendable rendering results using 3DGS, they still need to be improved because of the inherent inefficiency and discreteness of the 3DGS representations.  Furthermore, these approaches exclusively focus on modeling the frontal head, neglecting the rear and side view.

\subsection{3D-aware Generative Face Model}
Another avenue of research shifts the focus away from training person-specific avatars, instead emphasizing training a general facial prior with large-scale image datasets. Some of these studies~\cite{hong2021headnerf, buehler2021varitex, kirschstein2023nersemble, MoRF, AVA, buhler2023preface,zhuang2024nativegenerativemodel3d, Yan2024, wu2023describe3d} aim to construct a conditional model, utilizing expensive dense multi-view cameras or multi-view data obtained through light field capture to create rich conditions (\textit{e.g.}, identity, expression, direction). 
NeRSemble~\cite{kirschstein2023nersemble} constructs a multi-view radiance field to represent the human head, while AVA~\cite{AVA} develops a Gaussian variational autoencoder. 
MoFaNeRF~\cite{zhuang2022mofanerf} further introduces a refined GAN to enhance performance. 
Other work~\cite{Chan2022EG3D, PanoHead, sun2023next3d, piGAN2021, Schwarz2020NEURIPS} trains 3D-aware GAN from large-scale 2D image datasets (\textit{e.g.,} FFHQ~\cite{Karras_2019_CVPR_StyleGAN}). 
EG3D~\cite{Chan2022EG3D} introduces a novel triplane representation to render high-fidelity 3D heads with multi-view consistency, but only the front of the head. 
Next3D~\cite{sun2023next3d} introduces FLAME coefficients as conditions on top of EG3D but still does not reveal a full-head avatar. 
PanoHead~\cite{PanoHead} solves the problem by disambiguating the triplane and designing a complex pose estimation pipeline.
SphereHead~\cite{li2024spherehead} introduces a triplane representation in spherical coordinates and incorporates additional side and rear view data to enhance performance.
\section{Method}
\label{method}

The entire pipeline is shown in Fig.~\ref{fig: pipeline}, we first introduce the overall monocular reconstruction methods in Sec.~\ref{sec: overall}, then explain the sampling-based densification in Sec.~\ref{sec: densify}.  The neural baking, an optional module supporting texture-based editing, will be explained in Sec.~\ref{sec: baking}, and the universal completion framework to synthesize a 360$^\circ$-renderable head will be detailed in Sec.~\ref{sec: aug}.

\subsection{Monocular Reconstruction}
\label{sec: overall}

Following 3D Gaussian Splating~\cite{kerbl3Dgaussians}, our 3D head avatar is represented by $N$ unordered Gaussians $\mathcal{G} _i$, each of which possesses its own attributes:
\begin{align}
    \mathcal{G} _i=\left\{ \mathbf{p }_i,\boldsymbol{r}_i,\boldsymbol{s}_i,o_i,\boldsymbol{c}_i,d_i \right\},
\end{align}
\noindent where $\mathbf{p }_i$ is the Gaussian position in UV space, $\boldsymbol{r}_i$ and $\boldsymbol{s}_i$ is rotation vector and scaling vector to construct the covariance matrix, $o_i$ and $\boldsymbol{c}_i$ represent opacity and color respectively, and $d_i$ is the offset along the mesh normal. $\boldsymbol{r}_i$ and $\boldsymbol{s}_i$ represent local rotation and scaling. Given the rotation $\boldsymbol{R}$ and scale factor $k$ of mesh face, the global rotation $\boldsymbol{r}^{\prime}_i$ and $\boldsymbol{s}^{\prime}_i$ are expressed as:
\begin{align}
\boldsymbol{r}_{i}^{\prime}=\boldsymbol{Rr}_i,
\\
\boldsymbol{s}_{i}^{\prime}=k\boldsymbol{s}_i.
\end{align}
We sample uniformly in UV space to obtain $\mathbf{p}$, where each valid sample provides a set of barycentric coordinates $\left\{ \boldsymbol{w}_0,\boldsymbol{w}_1,\boldsymbol{w}_2 \right\} $ and a face index $f$. By the predefined UV mapping $\mathcal{M}(\cdot)$, $\mathbf{p}$ can be transformed into the 3D world coordinate. The offset $d$ is introduced along the normal direction $\mathbf{n}_f$. The Gaussian position can be formulated as:
\begin{align}
    \boldsymbol{\mu }=\mathcal{M} \left( \mathbf{p} \right) +d\cdot \mathbf{n}_f.
\end{align}
With such a formulation, the Gaussian position can move with the template mesh under various expressions and poses. Considering that the template mesh still differs significantly from the geometry in monocular video, we follow prior works~\cite{zhao2024psavatar, wu2024gaussianheadshoulders} to introduce personalized expression and pose blendshapes to model geometric gap:
\begin{align}
\mathbf{T}=\mathrm{LBS}\left( B_P\left( \Theta ;\mathcal{P} +\Delta \mathcal{P} \right) +B_E\left( \varPsi ;\mathcal{E} +\Delta \mathcal{E} \right) \right) ,
\end{align}
\noindent where $\mathbf{T}$ is the mesh with pose $\Theta$ and expression $\varPsi$, $\Delta \mathcal{E}$ and $\Delta \mathcal{P}$ are learnable blendshapes introduced, $\mathrm{LBS}(\cdot)$ denote the linear blendshape skinning function, as defined in~\cite{FLAME}. We observed that directly optimizing blendshapes leads to unstable and noisy mesh. Therefore, we introduce regularization terms on the mesh (See in Sec:~\ref{sec:mesh_loss}).

\begin{figure}[]
    \centering
    \includegraphics[width=0.5\textwidth]{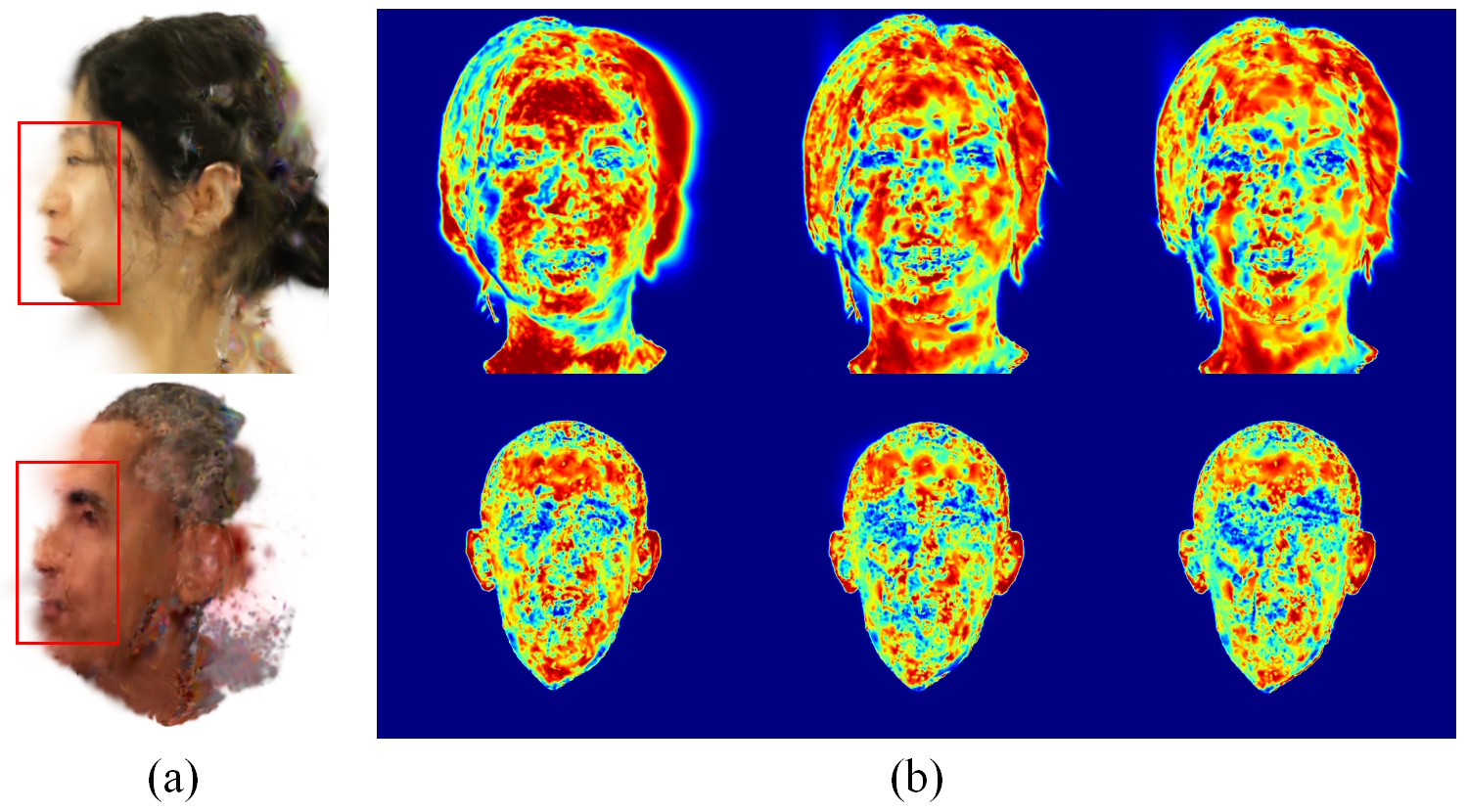}
    \vspace{-0.2in}
    \caption{\textbf{3DGS in Monocular Video}. (a) In monocular reconstruction, since the sides of the head avatar are rarely supervised, Gaussians tend to grow towards the direction of the rendering camera.
    (b) This potentially results in position gradient visualizations during training, showing that most of the facial region displays distributions exceeding the threshold $\tau_{\mathrm{pos}}$.}
    \label{fig: sampling_demo}
    \vspace{-10px}
\end{figure}

\subsection{Sampling-based Densification}
\label{sec: densify}

In the vanilla 3DGS, densification is performed by introducing position gradients $\left\| \frac{\partial \mathcal{L}}{\partial \boldsymbol{\mu }} \right\| $ as an effective performance metric. By setting a threshold $\tau_{\mathrm{pos}}$, Gaussians with gradients exceeding this threshold are cloned and splited~\cite{kerbl3Dgaussians}.  This threshold-based densification has two main limitations. Firstly, in UV space, Gaussian is defined by its face index and barycentric coordinates, restricting its mobility compared to that in view space. Secondly, threshold-based densification makes it challenging to control the Gaussian number, resulting in excessive Gaussian usage. 
It is worth noting that the predominance of frontal camera views in most monocular videos exacerbates this issue.  As shown in Fig.\ref{fig: sampling_demo} (b), we observed that a substantial number of Gaussians (\textit{e.g.}, cheek, forehead) appear to require frequent but unreasonable splits and clones, leading to redundancy in Gaussian numbers and imprecision in volumetric representation.  We believe this issue is unavoidable because it stems from the inherent ambiguity of monocular head pose estimation.

To solve this problem, we propose sampling-based densification. We retain $\left\| \frac{\partial \mathcal{L}}{\partial \boldsymbol{\mu }} \right\| $ as the performance metric. Instead of selecting a threshold $\tau_{\mathrm{pos}}$, we treat each Gaussian primitive $\mathcal{G}_i$ as proposal for their binding face $f_i$ and use $\left\| \frac{\partial \mathcal{L}}{\partial \boldsymbol{\mu }} \right\| $ as an importance metric $\mathcal{I}$ for multinomial sampling, with the probability that \(k\)-th Gaussian is selected as:
\begin{equation}
    p_k=\frac{\mathcal{I} _k}{\sum_{i=0}^{N-1}{\mathcal{I} _i}},
\end{equation}
\noindent where $N$ is the total number of Gaussian primitives. When the $k$-th Gaussian is selected, we can query the face index $f_i$ of the $k$-th Gaussian.  A set of barycentric coordinates in triangle $f_i$ is initialized as follows:
\begin{align}
    \boldsymbol{w}_j=\frac{\hat{\boldsymbol{w}}_j}{\sum_{m=0}^2{\hat{\boldsymbol{w}}_m}},\quad j=0,1,2,
    \\
    \hat{\boldsymbol{w}}_0,\hat{\boldsymbol{w}}_1,\hat{\boldsymbol{w}}_2\sim \mathcal{U}\left( 0,1 \right) .
\end{align}
In this way, a new Gaussian position is obtained. By letting the new Gaussian inherit the sampled splat's attributes, we achieve densification via a sampling approach. In the training phase, the densification is performed at regular intervals to sample a fixed number of Gaussians. Afterward, some unsuitable Gaussians will be pruned in the subsequent training iterations based on opacity conditions.  This prevents an explosion in the number of points while also allowing the distribution of Gaussians to update gradually in a controlled manner.

\begin{figure}[]
    \centering
    \includegraphics[width=0.45\textwidth]{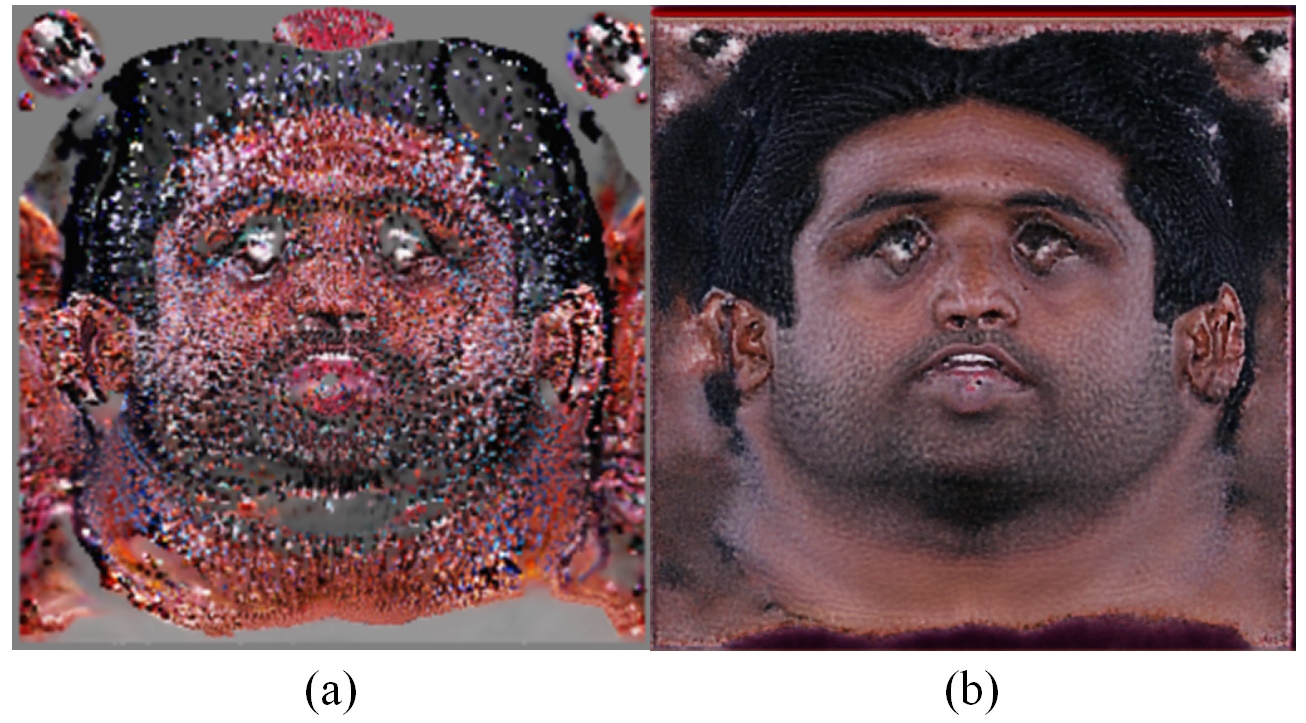}
    \vspace{-0.1in}
    \caption{\textbf{Texture Map Visualization}. (a) Directly optimizing texture maps often results in significantly low quality, with visible holes and artifacts. (b) In contrast, our neural baking method produces a much smoother and more plausible texture map.}
    \label{fig: bake_demo}
    \vspace{-5px}
\end{figure}

\subsection{Neural Baking for Texture Editing}
\label{sec: baking}

After learning an animatable Gaussian avatar with sample-based densification and optional full-head completion, we further propose the neural baking to edit the discrete 3D Gaussian avatar explicitly (Stage II in Fig.~\ref{fig: pipeline}).  Neural baking is defined as a process of transforming a discrete and unordered Gaussian attribute map into a continuous and editable one.  The specific implementation is achieved by introducing BakeNet for a two-stage training.

The raw learned Gaussian model is a discrete representation that is highly convenient for rendering, but the discrete and unordered point set is complicated to edit.
Since we have parameterized the Gaussians into 2D UV space, an intuitive idea is to construct a reconstruction kernel $\mathcal{R}(\cdot)$ that samples a continuous and smooth function $f(\cdot)$ from the discrete Gaussian attributes $w$:
\begin{align}
    f\left( \mathbf{p} \right) &=\left( w\ast \mathcal{R} \right) \left( \mathbf{p} \right) 
    \\
    &=\sum_k{w_k\mathcal{R} _k\left( \mathbf{p}-\mathbf{p}_k \right)}
\end{align}
\noindent where $\mathbf{p}$ is the UV coordinate. Directly constructing $\mathcal{R}(\cdot)$ is both manual and complex, as the properties and ranges of interpolation functions may vary across different Gaussian attributes. Considering that $\mathcal{R}(\cdot)$ only requires to satisfy \textit{local support}, we can select the bilinear interpolation operator $\mathcal{B}(\cdot)$ as the kernel and then focus on refining $w_k$ to ensure smoothness in $f(\cdot)$. Thus, our objective becomes finding a suitable proxy $\phi_k$ for Gaussian attributes $w_k$.

A straightforward solution to this objective is to approximate $\phi_k$ with $w_k$ by optimizing randomly initialized feature maps $\mathcal{F}$ and applying $\mathcal{B}(\cdot)$ over UV coordinates. However, experiments show that the result texture maps are discontinuous and messy, as shown in Fig.~\ref{fig: bake_demo} (a). We observed that such an issue doesn't exist in several generative Gaussian head models~\cite{kirschstein2024gghead,zhuang2024nativegenerativemodel3d,URAvatar,zhuang2024idolinstant}, of which the Gaussian attribute maps are continuous. We consider this phenomenon attributable to the inherent regularization properties of the convolutional operations incorporated into the generative model.  On further analysis, we argue that the inductive biases of the CNN contribute to local smoothness and translation invariance, serving as a pre-filter $\mathcal{H}(\cdot)$. Hence, $f\left( \mathbf{p} \right)$ can finally be formalized as:
\begin{align}
    f\left( \mathbf{p} \right) =\left( \mathcal{F} \ast \mathcal{H} \ast \mathcal{B} \right) \left( \mathbf{p} \right), 
\end{align}
where the low-pass $\mathcal{H}(\cdot)$ and $\mathcal{B}(\cdot)$ ensure the continuity of $f\left( \mathbf{p} \right)$.
Under the guidance of this idea, BakeNet is introduced as the pre-filter $\mathcal{H}(\cdot)$, which takes multi-channel noise maps sampled from Gaussian distribution $\mathcal{F}$ as input to regularize the attribute map in a post-training stage. U-Net~\cite{UNet} is selected as the backbone of the BakeNet.

The parameters of the BakeNet are updated by the gradients computed by the loss defined in Sec.~\ref{sec: loss}. 
We sample attributes from the U-Net output to replace the point-wise Gaussian attributes, inheriting the trained $\Delta \mathcal{E}, \Delta \mathcal{P}$ and sampled UV coordinates. After neural baking, the rendering quality may experience degradation. The BakeNet will not be involved in model inference but only help regularize the attribute maps in the stage II training.  Experiments demonstrate that this two-stage learning strategy leads to higher rendering performance and faster convergence speed than direct end-to-end training with BakeNet. We also study to improve the rendering quality of the baked results and further discuss the trade-off between rendering quality and texture quality. Due to space constraints, these are placed in the supplementary reporting material.

\begin{figure}[t]
    \centering
    \includegraphics[width=0.45\textwidth]{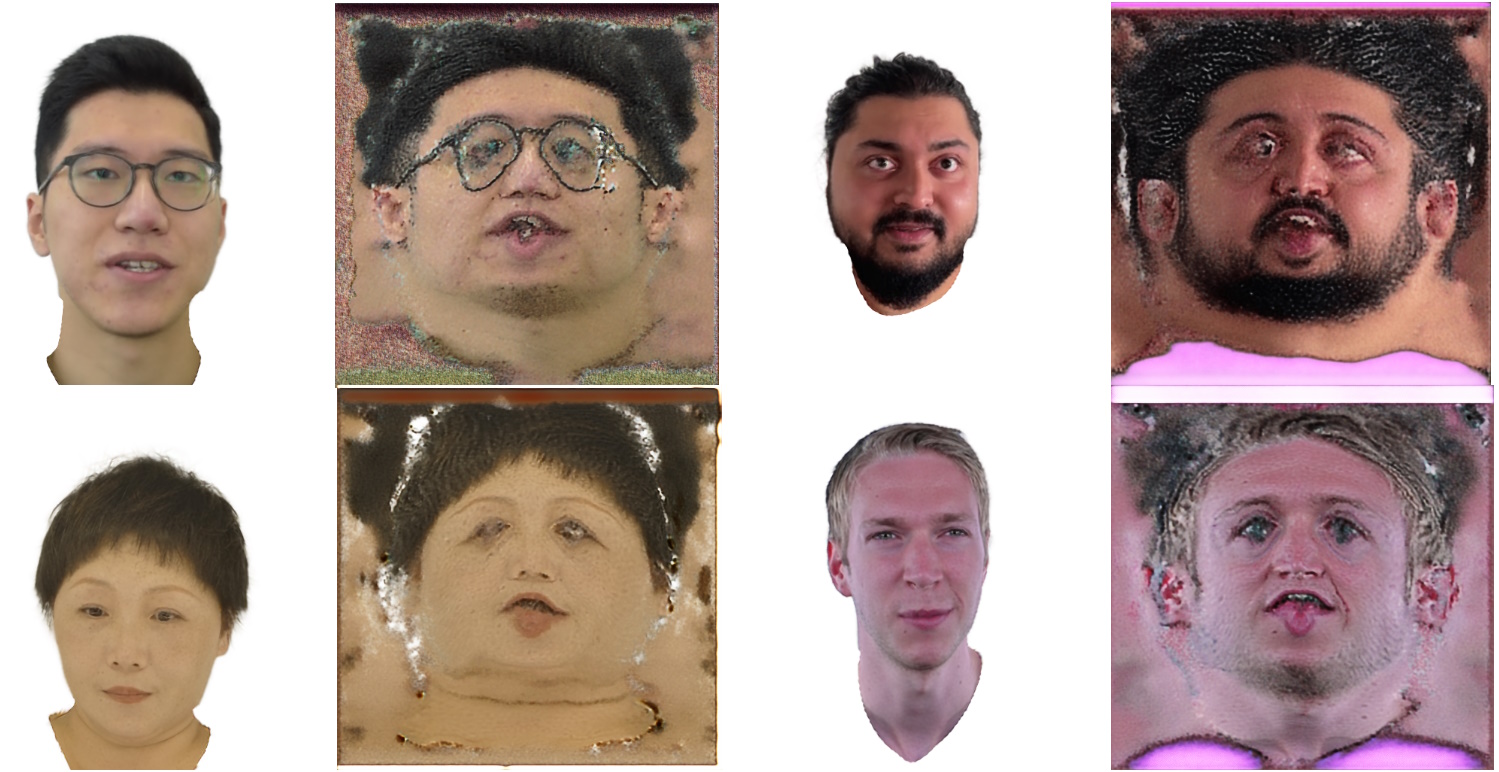}
    \vspace{-0.1in}
    \caption{\textbf{Baked Results Visualization}. We visualize the color texture map produced by neural baking on different subjects.
    }
    \label{fig: bake_result}
    \vspace{-15px}
\end{figure}

\subsection{Full-Head Completion}
\label{sec: aug}

\begin{figure*}[!htbp]
    \centering
    \includegraphics[width=\textwidth]{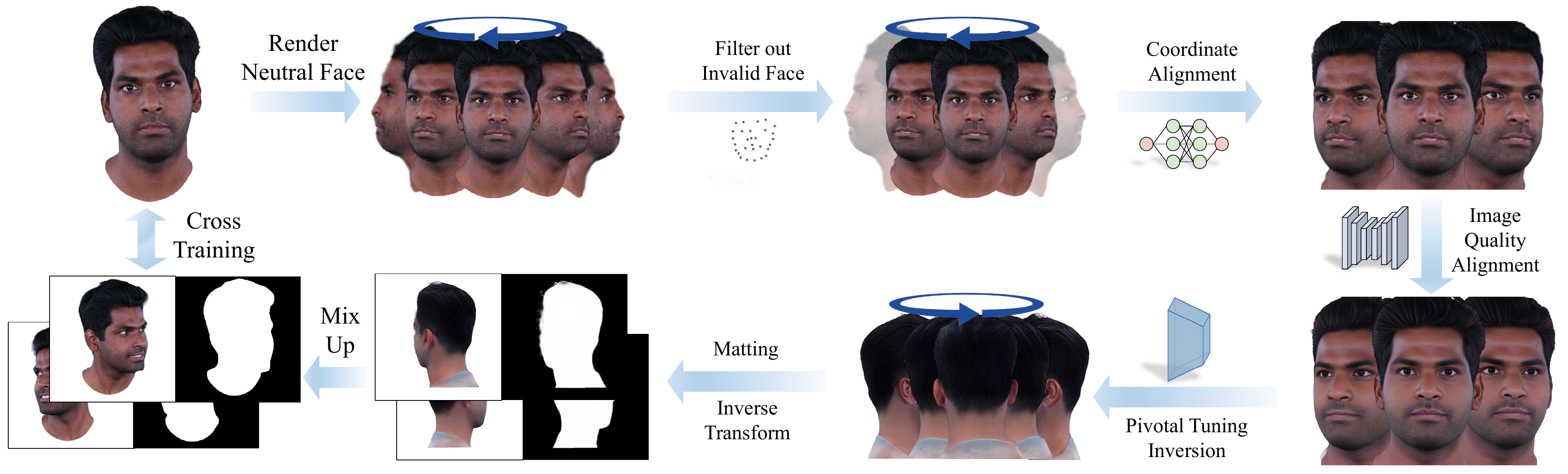}
    \vspace{-0.2in}
    \caption{\textbf{Completion Framework}. A universal framework is proposed to complete the side and rear appearance under monocular settings. 
    }
    \label{fig: full_head_framework}
    \vspace{-15px}
\end{figure*}

Previous monocular head reconstruction algorithms have typically neglected hair modeling for two primary reasons.  Firstly, the rear region of the head is commonly featureless hair, where pose tracking and 3DMM regression always fail.  Secondly, most portrait videos focus on the frontal face, with no specific capture of the rear head.  For these reasons, an intuitive solution is to leverage pretrained full-head generative models~\cite{li2024spherehead, PanoHead} to synthesize rear head frames.

However, generating images to reconstruct the rear head appearance is nontrivial.  Existing full-head generative models set up a canonical model space with simplified orthogonal projection, which differs from monocular video-based reconstruction. Therefore, establishing model space transformation and enhancing the quality of rear head generation become the most critical issues. To solve these issues, we design a universal completion framework by extracting priors from SphereHead~\cite{li2024spherehead} for completing the rear head of the learned animatable head avatar. The proposed completion framework consists of three steps: coordinate alignment, image quality alignment, inversion and finetuning.

\begin{figure*}[!htbp]
    \centering
    \includegraphics[width=0.95\textwidth]{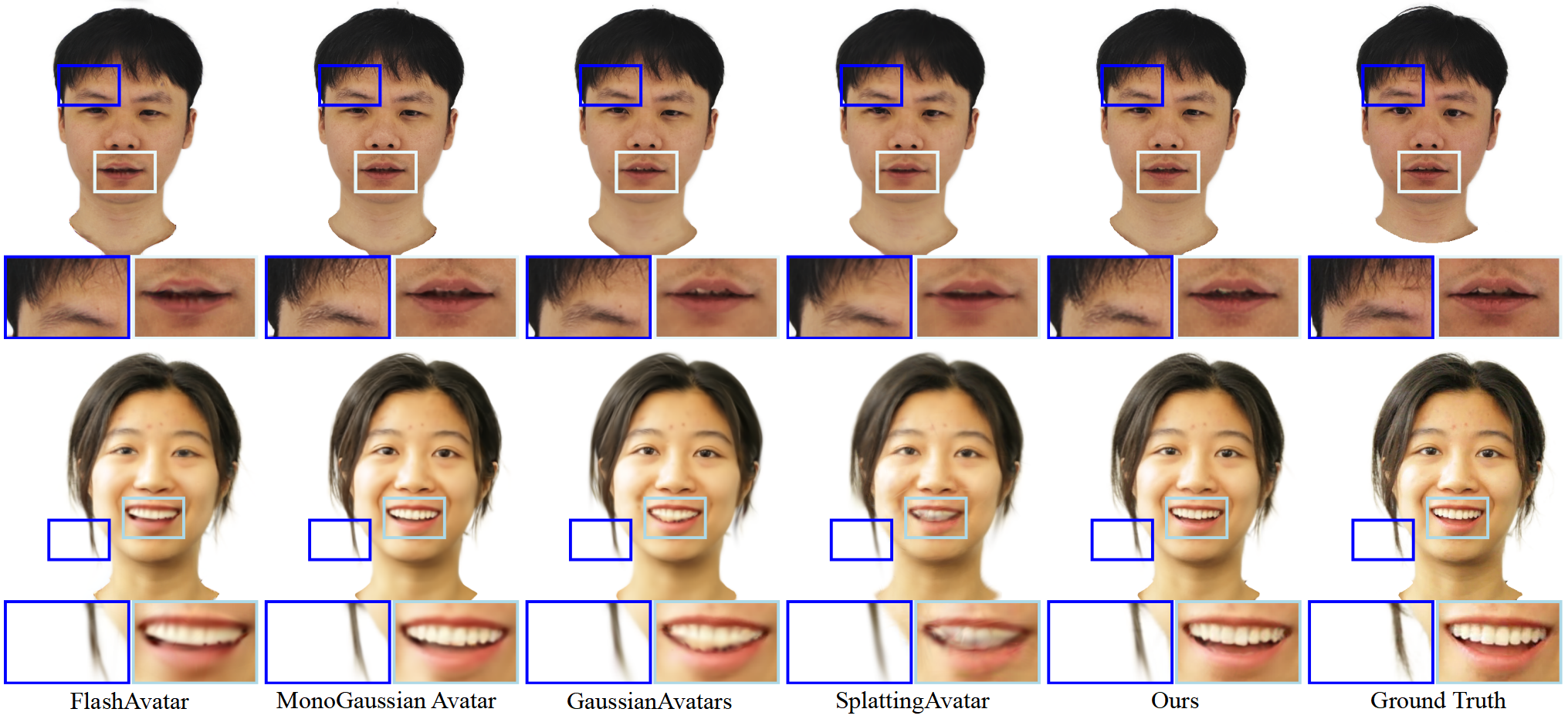}
    \vspace{-1em}
    \caption{\textbf{Monocular Reconstruction Results}. Our method is more effective at capturing fine structure and high-frequency details (\textit{e.g.} loose strands of hair, lip creases, and stubble in the facial area.). More reconstructed subjects are shown in supplementary materials.
    }
    \label{fig: mono_recon}
    \vspace{-8px}
\end{figure*}

\begin{figure*}[!htbp]
    \centering
    \includegraphics[width=0.96\textwidth]{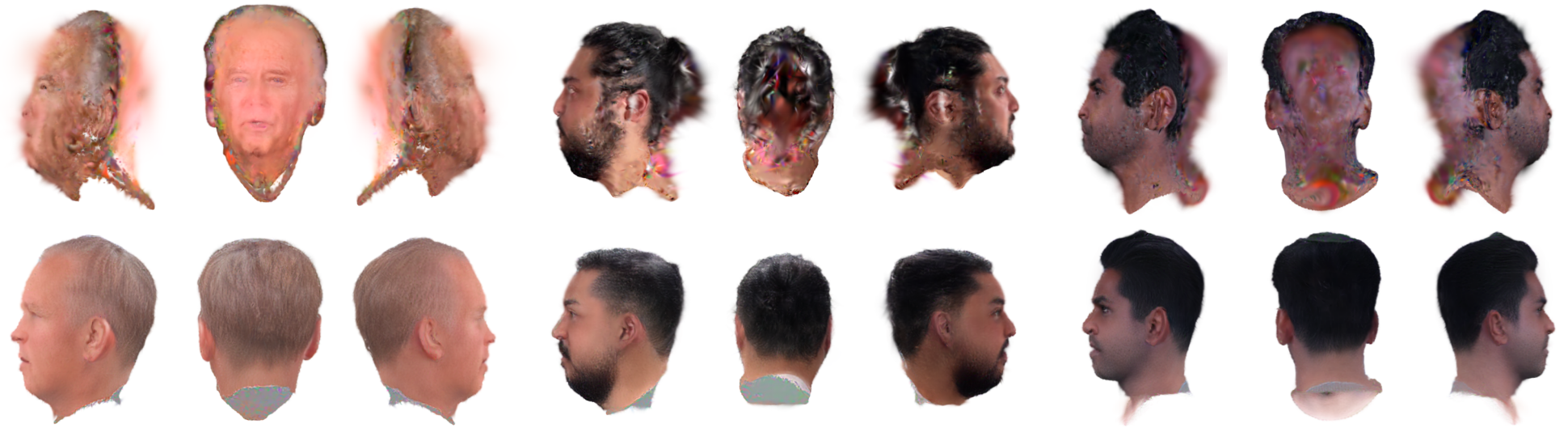}
    \vspace{-1em}
    \caption{\textbf{Full-head Completion Results}. The first row shows the side and back views rendered in our method without completion, and the second row shows the result after completion. 
    }
    \label{fig: full_head}
    \vspace{-10px}
\end{figure*}

\begin{figure*}[!htbp]
    \centering
    \includegraphics[width=0.95\textwidth]{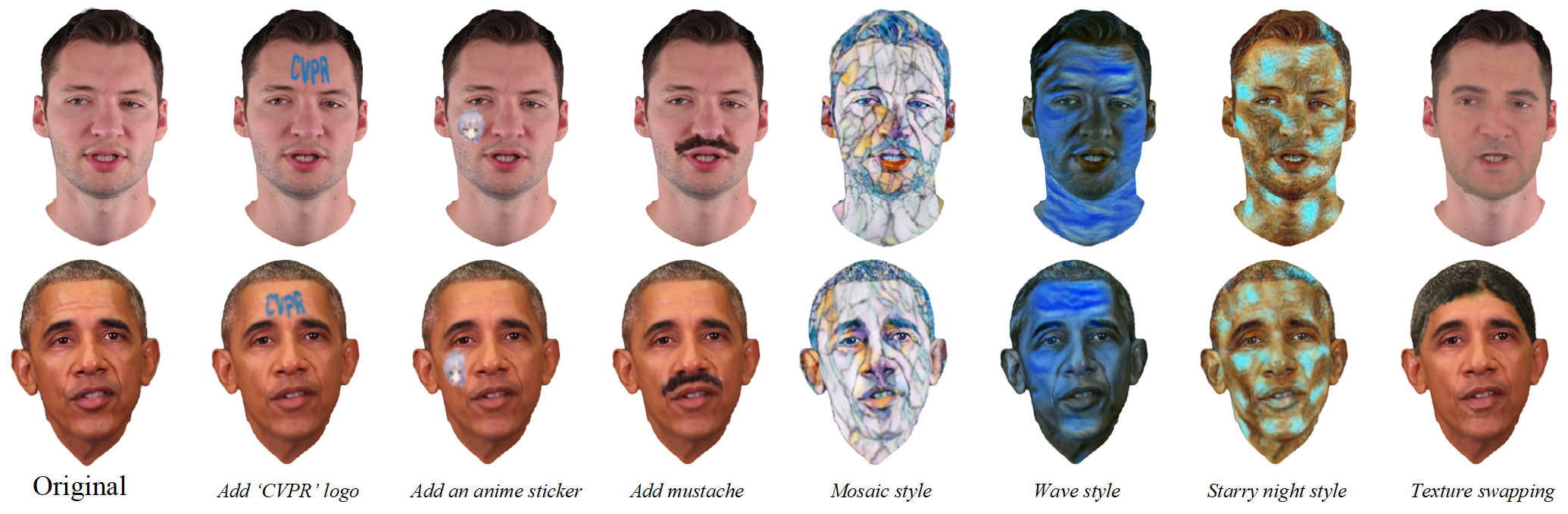}
    \vspace{-1em}
    \caption{\textbf{Texture Editing Results}. We show the effects of simply and effectively editing the baked texture map.
    }
    \label{fig: edit_result}
\end{figure*}

\noindent \textbf{Coordinate alignment.}
First, we set up a horizontal circle of camera orbit to render around the head avatar with neutral expression and pose. Choosing neutral expression and pose is because SphereHead excels at representing static faces, and neutral status simplifies subsequent alignment and inverse transformations. Then, a face detector~\cite{dlib} is used to assess landmark confidence in all rendered views and filter out the side-view images with low confidence scores. We employ TDDFA~\cite{guo2020towards} to obtain facial keypoints $\mathbf{Q}=\left[ \mathbf{q}_1,...,\mathbf{q}_{68} \right] \in \mathbb{R} ^{2\times 68}
$. $\mathbf{Q}$ is used to construct an affine transformation matrix $\mathcal{A}$ for image cropping and aligning.

\noindent \textbf{Image quality alignment.}
Directly using the rendered aligned images for Pivotal Tuning Inversion (PTI)~\cite{pti} often produces blurry results. We consider the reason to be the domain gap between the image quality of the input video and the high-quality dataset used to train SphereHead.  Therefore, we utilize a face restoration model, GFPGAN~\cite{gfpgan}, to align the image quality of the input video and SphereHead. As GFPGAN is trained on a data source similar to the SphereHead dataset, it can inject image quality-level details into the input video frames, helping fit the distribution of SphereHead training set.
As our primary goal is to leverage the priors from SphereHead regarding side and rear views, some identity changes caused by GFPGAN in the frontal view are acceptable.

\noindent \textbf{Inversion and finetuning.}
We extend PTI to multiple images, using valid multi-view faces filtered by the aforementioned facial landmark detector for supervision. For a detailed formulation of the optimization process, please refer to the supplementary materials. 
After obtaining the inverted orbited images, we utilize the estimated facial landmarks $\mathbf{Q}$ to calculate an affine transformation matrix $\mathcal{A}^{-1}$ using the least squares optimization. $\mathcal{A}^{-1}$ is applied to transform all synthesized images. Then, MODNet~\cite{MODNet} is used to extract facial masks of the synthesized images. We cross-train from these pseudo-images and ground truth to avoid degradation of the frontal view.

\subsection{Training Objective}
\label{sec: loss}

The optimization goal is to minimize the loss between the rendered output and the ground truth, subject to certain regularization constraints. The first term is the image loss:
\begin{equation}
    \mathcal{L} _{\mathrm{image}}=\mathcal{L} _{\mathrm{L}1}+\lambda _1\mathcal{L} _{\mathrm{vgg}}.
\end{equation}
To avoid Gaussians becoming over-skinny, we introduce the regularization term following PhysGaussian~\cite{xie2023physgaussian}:
\begin{equation}
    \mathcal{L} _{\mathrm{scale}}=\frac{1}{N}\sum_{i=0}^{N-1}{\max \left( \frac{\max \left( \boldsymbol{s}_i \right)}{\min \left( \boldsymbol{s}_i \right)}-r,0 \right)},
\end{equation}
\noindent where $N$ is the total number of splats, and $r$ is a hyperparameter. This loss ensures that the ratio of major axis length to minor axis length stays below $r$. 
\label{sec:mesh_loss}
Moreover, we employ additional regularization terms specific to the mesh to constrain its geometry:
\begin{equation}
    \mathcal{L} _{\mathrm{mesh}}=\lambda _2\mathcal{L} _{\mathrm{lap}}+\lambda _3\mathcal{L} _{\mathrm{flame}},
\end{equation}
\noindent where $\mathcal{L} _{\mathrm{lap}}$ is the laplacian smoothing term, $\mathcal{L} _{\mathrm{flame}}$ is $\mathrm{L}2$ distance between current vertices and original vertices in given pose and expression.

The overall loss function is defined as:
\begin{equation}
\mathcal{L} =\mathcal{L} _{\mathrm{L}1}+\lambda _1\mathcal{L} _{\mathrm{vgg}}+\lambda _2\mathcal{L} _{\mathrm{lap}}+\lambda _3\mathcal{L} _{\mathrm{flame}}+\lambda _4\mathcal{L} _{\mathrm{scale}},
\end{equation}
\noindent where $\lambda_1, \lambda_2, \lambda_3, \lambda_4$ are set to $0.1$, $100$, $100$, $0.1$.

\begin{table*}[!htbp]
\caption{Comparison of quantitative results with state-of-the-art methods. \colorbox{blue!50} {blue} and \colorbox{blue!20}{lightblue} indicate the 1st and 2nd best.}
\vspace{-0.1in}
\centering
\small
\scalebox{0.73}{
\begin{tabular}{c|ccccccccccccccc}
\toprule
\multirow{2}{*}{Datasets} & \multicolumn{3}{c|}{Overall}                & \multicolumn{3}{c|}{INSTA Dataset} & \multicolumn{3}{c|}{PointAvatar Dataset} & \multicolumn{3}{c|}{NerFace Dataset} & \multicolumn{3}{c}{Ours Dataset} \\ \cline{2-16} 
                          & PSNR↑ & SSIM↑ & \multicolumn{1}{c|}{LPIPS↓} & PSNR↑     & SSIM↑      & LPIPS↓    & PSNR↑      & SSIM↑      & LPIPS↓      & PSNR↑      & SSIM↑      & LPIPS↓     & PSNR↑    & SSIM↑     & LPIPS↓    \\ \midrule
FA~\cite{xiang2024flashavatar} (\textit{CVPR'24})                                   & 27.41      & 0.9322      & \clb{0.0603}                            & 27.28     & 0.9346     & \cb{0.0578}    & 26.45      & 0.9103      & \clb{0.0890}             & 31.38      & 0.9641     & \clb{0.0304}     & \clb{25.48}    & 0.9188    & \cb{0.0679}    \\
SA~\cite{SplattingAvatar:CVPR2024} (\textit{CVPR'24})                                   & 26.34      & 0.9249      & 0.1135                            & 26.63     & 0.9304     & 0.1119    & 24.76      & 0.8907      & 0.1501             & 29.34      & 0.9480     & 0.0712     & 24.55    & 0.9196    & 0.1218    \\
MGA~\cite{chen2024monogaussianavatar} \textit{\fontsize{7}{11}\selectfont (SIGGRAPH'24)}  & \clb{28.07}      & \clb{0.9405}      & 0.0787                            & \clb{27.40}     & 0.9373     & 0.0887    & \clb{28.16}      & \cb{0.9360}      & 0.0977             & \cb{33.78}      & \cb{0.9765}     & 0.0315     & 25.43    & 0.9210    & 0.0749    \\
GA~\cite{qian2023gaussianavatars} (\textit{CVPR'24})                                   & 26.20      & 0.9343      & 0.0804                            & 26.66     & \clb{0.9396}     & 0.0777    & 24.51      & 0.9078      & 0.1257             & 29.06      & 0.9559     & 0.0509     & 24.18    & \clb{0.9251}    & 0.0755    \\
Ours                                                    & \cb{28.37}      & \cb{0.9439}      & \cb{0.0586}             & \cb{27.52}     & \cb{0.9416}     & \clb{0.0603}    & \cb{28.74}      & \clb{0.9333}      & \cb{0.0719}             & \clb{33.70}      & \clb{0.9736}     & \cb{0.0257}     & \cb{26.25}    & \cb{0.9358}    & \clb{0.0691}    \\ \hline
Ours (baked)                                            & 28.23      & 0.9415      & 0.0676             & 27.80     & 0.9419     & 0.0639    & 27.45      & 0.9239      & 0.0954             & 32.59      & 0.9665     & 0.0373     & 26.13    & 0.9326    & 0.0823    \\ \bottomrule
\end{tabular}
}
\vspace{-12px}
\label{tab:mono_recon}
\end{table*}

\begin{table}[!htbp]
\centering
\caption{Comparison of the number of Gaussians}
\vspace{-0.15in}
\small
\scalebox{0.9}{
\begin{tabular}{ccccc}
\toprule
Data & INSTA & IMAvatar & NerFace & EmoTalk3D  \\ \hline
FA       & \multicolumn{4}{c}{-------------------------\quad16k\quad--------------------------}  \\
SA       & 558k±188k & 617k±274k & 497k±142k & 489k±171k \\
MGA      & \multicolumn{4}{c}{-------------------------\quad100k\quad-------------------------}  \\
GA       & 72k±33k   & 38k±14k    & 31k±7k    & 55k±12k    \\
Ours     & 49k±6k    & 38k±6k    & 42k±0.5k  & 58k±2k    \\ \bottomrule
\end{tabular}
}
\label{tab:gs_number}
\vspace{-8px}
\end{table}

\begin{table}[]
\centering
\caption{Ablation Study in \textit{yufeng} case.}
\vspace{-0.1in}
\scalebox{0.8}{\begin{tabular}{llll}
\toprule
                   & PSNR↑ & SSIM↑  & LPIPS↓ \\ \hline
Ours               & 29.36 & 0.9239 & 0.0694 \\
w/o densify        & 29.13 & 0.9217 & 0.0740 \\
w/o $\Delta \mathcal{E}$ and $\Delta \mathcal{P}$ & 24.78 & 0.8820 & 0.1112 \\
\hline
Two-stage baking   & 27.78 & 0.9104 & 0.0979 \\ 
One-stage baking   & 27.42 & 0.9085 & 0.1088 \\
Decode only        & 25.56 & 0.8878 & 0.1506 \\ \bottomrule

\end{tabular}}
\label{tab: abl_1}
\vspace{-15px}
\end{table}
\section{Experiments}
\label{exp}

We conduct extensive experiments across various datasets. A total of 20 subjects from different datasets are collected - 10 subjects from INSTA~\cite{INSTA}, preprocessed by the MICA tracker~\cite{MICA:ECCV2022};
3 subjects from PointAvatar~\cite{Zheng2023pointavatar}; 3 subjects from NerFace~\cite{nerface} processed using a DECA-based pipeline~\cite{zheng2022imavatar}; and 4 subjects in Emotalk3D~\cite{he2024emotalk3d}, also preprocessed via the DECA. Four state-of-the-art GS-based reconstruction methods are compared, including GaussianAvatars (GA)~\cite{qian2023gaussianavatars}, FlashAvatar (FA)~\cite{xiang2024flashavatar}, MonoGaussianAvatar (MGA)~\cite{chen2024monogaussianavatar} and SplattingAvatar (SA)~\cite{SplattingAvatar:CVPR2024}.

\subsection{Implementation Details}
We uniformly sample 65k Gaussians in the UV space. Given the consistent lighting condition in monocular video, we use zero-degree SH to represent color. We increase 1k Gaussians every 3k iterations. All experiments are conducted on a single A6000 GPU. Please refer to the supplementary materials for further details.

\subsection{Monocular Results}
Average PSNR, SSIM, and LPIPS~\cite{lpips} are reported in Tab.~\ref{tab:mono_recon}. Our method achieves balance among these metrics, delivering the best overall performance. As shown in Fig.~\ref{fig: mono_recon}, our method more effectively captures the high-frequency details of avatars while avoiding the needle-like artifacts often observed in 3DGS. Tab.~\ref{tab:gs_number} presents the number of Gaussians each method utilizes. Our method employs a rather small number of Gaussian primitives, and the variance of the Gaussian number is more stable in different datasets. This demonstrates the effectiveness of sampling-based densification. For more results on computational efficiency, please refer to the appendix.

\subsection{Neural Baking Results}

We visualize color texture maps of several head avatars generated through neural baking in Fig.~\ref{fig: bake_result}. The resulting texture maps exhibit smooth and continuous qualities, with neural baking interpolating reasonable details in regions not visible in the monocular video. Such quality texture maps enable straightforward editing. In Fig.~\ref{fig: edit_result}, we demonstrate various editing operations. Unlike previous approaches, our method allows precise control without cumbersome optimization.

\subsection{Full-Head Completion Results}
We show the rendered results of monocular reconstructed head avatars from our method after passing through the completion framework in Fig.~\ref{fig: full_head}. The significant improvement in the side and rear views demonstrates the effectiveness of the completion framework. This pipeline can be naturally extended to other methods, and we present the completed results 4 baselines in the supplementary materials.

\subsection{Ablation Study}
Ablation study are conducted on several designs in monocular reconstruction and neural baking. Quantitative results can be found in Tab.~\ref{tab: abl_1}, more results in the appendix.

\noindent \textbf{(i) w/o densify} When sampling-based densification is disabled, the LPIPS is considerably degraded. This suggests that the initialized uniform distribution is suboptimal.

\noindent \textbf{(ii) w/o $\mathbf{\Delta }\mathcal{E} $ and $\mathbf{\Delta }\mathcal{P} $} We set learnable blendshapes as fixed zero vectors. Without making FLAME learnable, degraded results are produced based on the coarse template.

\noindent \textbf{(iii) One-stage baking v.s. two-stage baking.} One-stage baking is to train the BakeNet together with the Gaussians in a single stage. We have discovered that it notably affects training efficiency and results in inferior rendering quality.

\noindent \textbf{(iv) Decode only} We only use the decoder of BakeNet for neural baking. The degradation indicates the effectiveness of the BakeNet for encoding high-frequency input.

\section{Conclusion}
\label{sec:con}
We propose a novel monocular video reconstruction method with sampling-based densification and neural baking for efficient appearance editing in the UV space. And a universal completion framework improves non-frontal view reconstruction, enabling 360$^\circ$-renderable 3D head avatars.

Limitations remain. Our method assumes consistent and uniform lighting, reducing robustness in real-world scenarios. The completion framework depends on the pre-trained model’s dataset, limiting its ability to capture complex, personalized head shapes and potentially causing identity change. Fixed-size texture maps from neural baking may also fail in some cases, which could be mitigated by baking with a Mip-Map mechanism. Future work could explore integrating full-body priors, such as SMPL-X~\cite{SMPL-X}, to enhance immersive applications.

\section*{Acknowledgements}
This study was funded by NKRDC 2022YFF0902200 and NSFC 62472213. Jiawei Zhang would like to thank Prof. Zhixi Feng for his support in the early stages of this study.

{
    \small
    \bibliographystyle{ieeenat_fullname}
    \bibliography{main}
}

\clearpage
\clearpage
\setcounter{page}{1}
\maketitlesupplementary

This supplementary material provides additional implementation details and experimental results. In Sec.~\ref{supp: preliminary}, we introduce the preliminaries related to 3DGS and PTI. Sec.~\ref{supp: implement_details} describes implementation details regarding datasets, methods, neural baking and head completion. In Sec.~\ref{supp: additional_results}, we present additional experimental results, including monocular reconstruction, cross-reenactment, more results about full-head completion, and textural editing. Sec.~\ref{supp: trade-off} explains the trade-off between texture quality and rendering quality in neural baking. We discuss the failure cases and ethics considerations in Sec.~\ref{supp: failure} and Sec.~\ref{supp: ethics}, respectively.
We integrate the performance under imperfect poses, computational efficiency, and additional ablation in Sec.~\ref{supp: noise_pose}, Sec.\ref{supp: compute}, and Sec.~\ref{supp: more_abl}, respectively. We highly recommend watching our \textit{supplementary video} for more visual results.

\section{Preliminary}
\label{supp: preliminary}
\paragraph{3D Gaussian Splatting}
3D Gaussian Splatting~\cite{kerbl3Dgaussians} is a point-based volume rendering method that models each primitive as a Gaussian kernel, formalized as follows:
\begin{align}
    G\left( \mathbf{x} \right) =e^{-\frac{1}{2}\left( \mathbf{x}-\boldsymbol{\mu } \right) ^T\mathbf{\Sigma }^{-1}\left( \mathbf{x}-\boldsymbol{\mu } \right)},
\end{align}
where $\boldsymbol{\mu }$ is Gaussian position and $\mathbf{\Sigma }$ is 3D covariance matrix. To ensure that $\mathbf{\Sigma }$ is positive semi-definite, the covariance matrix is further decomposed into a rotation matrix $\mathbf{R}$ and a scaling matrix $\mathbf{S}$:
\begin{align}
    \mathbf{\Sigma }=\mathbf{RSS}^T\mathbf{R}^T.
\end{align}

In the rendering phase, 3D Gaussians are projected onto the image plane as 2D Gaussians. Zwicker~\etal~\cite{EWASplatting} derive the following formula to approximate the covariance of the projected 2D Gaussians:
\begin{align}
    \mathbf{\Sigma }^{\prime}=\mathbf{JW\Sigma J}^T\mathbf{W}^T,
\end{align}

\noindent where $\mathbf{W}$ is viewing transformation and $\mathbf{J}$ is the Jacobian of the affine approximation of the projective transformation. Volumetric rendering is then performed for each pixel to calculate the final color:

\begin{align}
    \mathbf{C}=\sum_{i\in N}{\mathbf{c}_i\alpha _i\prod_{j=1}^{i-1}{\left( 1-\alpha _j \right)}},
\end{align}

\noindent where $\mathbf{c}_i$ is the color of each Gaussian and $\alpha _i$ represents the density computed by the projected Gaussians with $\mathbf{\Sigma }^{\prime}$ multiplied by each Gaussian's opacity $o_i$.

\paragraph{Pivotal Tuning Inversion}
\label{supp:pti_opt}

We introduce the overall PTI~\cite{pti} optimization pipeline as follows. In the first stage, we search for the pivotal latent code $w_p$ by minimizing:

\begin{align}
    \underset{w}{\mathrm{arg}\min}&\sum_{i=0}^{M-1}{\mathcal{L} _{\mathrm{prec}}\left( \mathrm{I}_{i}^{\mathcal{M} _{\mathrm{R}}},\mathrm{I}_{i}^{\mathcal{G}} \right)},
    \\
    \mathrm{I}_{i}^{\mathcal{G}}&=\mathcal{G} _{\mathrm{P}}\left( w,c_i;\theta \right) ,
\end{align}

\noindent where $M$ is the number of valid multi-view images, $\mathcal{L} _{\mathrm{prec}}$ denotes the perceptual loss~\cite{johnson2016perceptual}, $\mathrm{I}^{\mathcal{M} _{\mathrm{R}}}$ is the face image restored by pretrained model $\mathcal{M}_\mathrm{R}$, $\mathcal{G} _{\mathrm{P}}$ is the freezed pretrained generator, $c$ is the camera pose.

In the second stage, we finetune the generator parameters by minimizing the following loss term:

\begin{align}
\mathcal{L} _{\mathrm{pt}}&=\sum_{i=0}^{M-1}{\mathcal{L} _{\mathrm{prec}}\left( \mathrm{I}_{i}^{\mathcal{M} _{\mathrm{R}}},\mathrm{I}_{i}^{\mathcal{G}} \right) +\lambda _{\mathrm{L}2}\mathcal{L} _{\mathrm{L}2}\left( \mathrm{I}_{i}^{\mathcal{M} _{\mathrm{R}}},\mathrm{I}_{i}^{\mathcal{G}} \right)},
\\
\mathrm{I}_{i}^{\mathcal{G}}&=\mathcal{G} _{\mathrm{P}}\left( w_p,c_i;\theta ^{\ast} \right),
\end{align}

\noindent where $\theta^\ast$ is the tuned weights initialized with the pre-trained weights $\theta$.

\section{Implementation Details}
\label{supp: implement_details}
\subsection{Datasets}
We used a total of 20 monocular portrait videos for our experiments. For 10 datasets with DECA-based preprocessing, we optimize the DECA-predicted FLAME coefficients during training and testing in line with IMAvatar~\cite{zheng2022imavatar}. For the test-time fine-tuning, we perform FLAME coefficients optimization for 50 epochs. We optimize the FLAME coefficients with a learning rate of $5\times 10^{-4}$. For all datasets, a pre-trained segmentation model~\cite{bisenet} is used to remove regions below the neck to facilitate comparison. All methods except MonoGaussianAvatar are trained for 10 epochs on the INSTA and Emotalk3D datasets and 50 epochs on the PointAvatar and NerFace datasets.

\subsection{Models}
All methods are implemented by PyTorch~\cite{pytorch} with differential Gaussian rasterization from 3DGS~\cite{kerbl3Dgaussians}. And all methods are optimized by Adam~\cite{kingma2017adam} optimizer. To model the mouth region, each method incorporates the FLAME template with additional faces to close the mouth cavity, similar to FlashAvatar~\cite{xiang2024flashavatar}.

\noindent \textbf{Ours}
For our method, the learning rates for color, opacity, scale, rotation, and offset are \(2.5 \times 10^{-3}\), \(5.0 \times 10^{-2}\), \(5.0 \times 10^{-3}\), \(1.0 \times 10^{-3}\), and \(1.6 \times 10^{-3}\), respectively. The learning rate for the learnable blendshapes is \(1.0 \times 10^{-5}\). The opacity of the Gaussians is reset every $6k$ iterations, and sampling-based densification is performed every $3k$ iterations by adding $1k$ Gaussians. Pruning is conducted every $2k$ iterations based on an opacity threshold of \(5.0 \times 10^{-3}\).

\noindent \textbf{FlashAvatar}
FlashAvatar maintains a fixed number of Gaussians in the canonical space and utilizes an MLP-based deformer to learn the offset of scale, rotation, and position. And  We set the learning rate for the deformer to \(1.0 \times 10^{-4}\), for color to \(2.5 \times 10^{-3}\), for opacity to \(5.0 \times 10^{-2}\), for scale to \(5.0 \times 10^{-3}\), and for rotation to \(1.0 \times 10^{-3}\). The deformer has a hidden dimension of 256 and an output dimension of 10. The output channel corresponding to rotation is activated using an exponential function to ensure non-negativity. The scale offset, after being activated by the exponential function, is applied multiplicatively to the original unactivated Gaussian scale. At initialization, we perform uniform UV sampling at a resolution of 128. In addition to the uniform sampling, we apply additional random sampling, resulting in a total of $16k$ Gaussians.

\noindent \textbf{GaussianAvatars}
GaussianAvatars was originally designed for multi-view video datasets with accurate 3D mesh, whereas the preprocessing pipeline for monocular videos cannot obtain such precise geometry prior. Due to its specific binding mechanism, we set the learning rate for scale to \(1.7 \times 10^{-2}\). Densification starts after $10k$ iterations and is performed every $2k$ iterations thereafter. The densification gradient threshold is \(1.0 \times 10^{-4}\), and Gaussians are pruned with a minimum opacity threshold of \(5.0 \times 10^{-3}\).

\noindent \textbf{MonoGaussianAvatar}
MonoGaussianAvatar employs a series of MLPs to model geometry, deformation, and Gaussian attributes. The design of the MLPs follows the original implementation, with a learning rate of \(1.0 \times 10^{-4}\). Densification of Gaussians is performed on an epoch-based scheduler, and the scheduler for the number of Gaussians added during densification remains consistent with the original paper. We perform densification every 5 epochs. Due to the slow convergence of MonoGaussianAvatar, we train each subject for 100 epochs.

\noindent \textbf{SplattingAvatar}
SplattingAvatar constructs Gaussians that walk on triangles using UV coordinates. We set the learning rate for UV coordinates (and the normal offset \(d\)) to \(1.6 \times 10^{-4}\), while the learning rates for other attributes remain consistent with the original Gaussian configuration. Opacity is reset every $3.5k$ iterations, and the walking triangle operation is performed every 100 iterations. The densification gradient threshold is set to \(2.0 \times 10^{-4}\), and the minimum opacity for pruning is \(5.0 \times 10^{-3}\). During initialization, we sampled $10k$ Gaussian points.

\begin{figure}[t]
    \centering
    \includegraphics[width=0.5\textwidth]{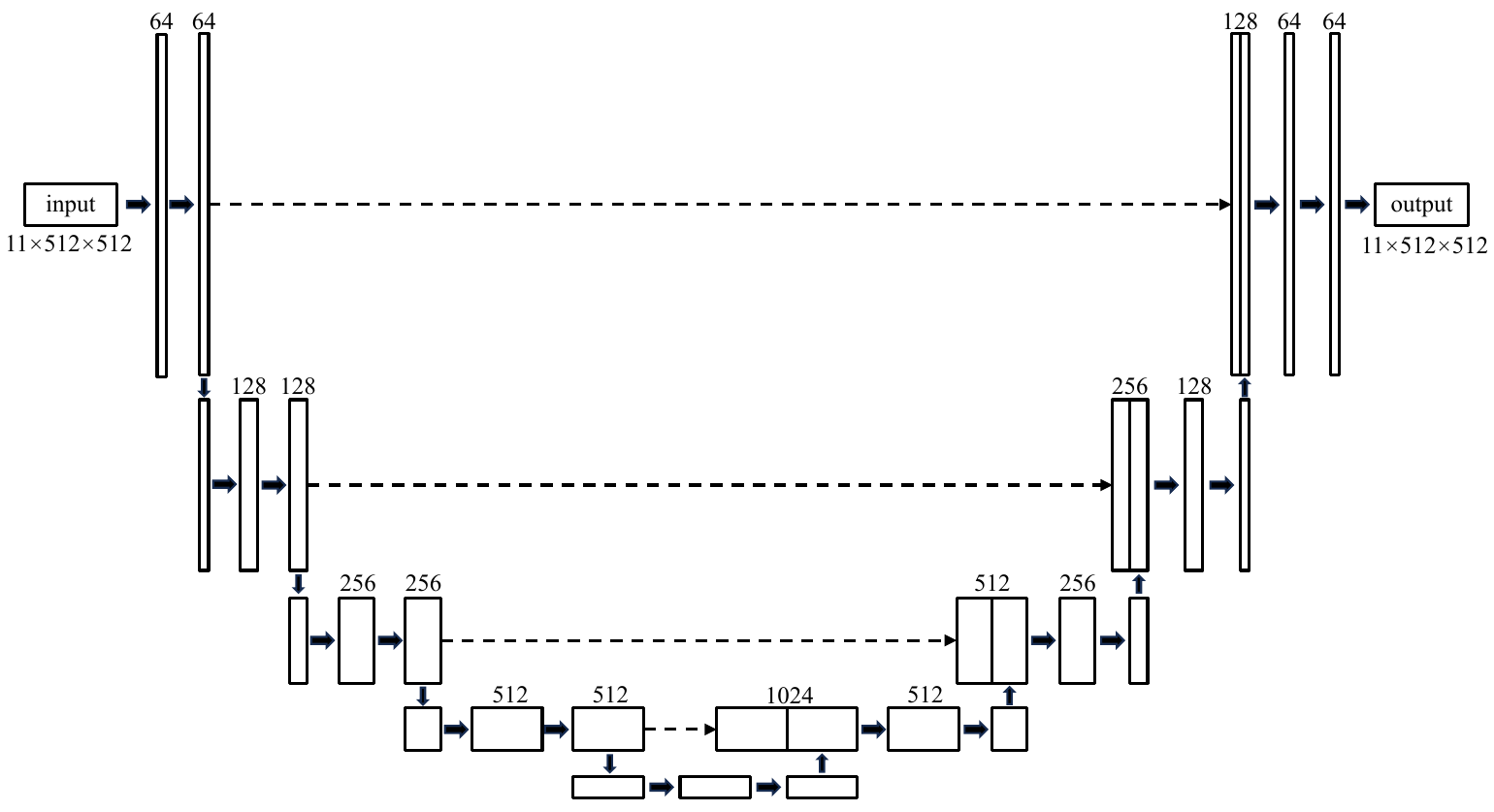}
    \caption{\textbf{BakeNet Architecture}.
    We adopt a U-Net architecture as the backbone of BakeNet, leveraging its ability to construct representations across various frequency bands from noise.
    }
\label{fig: unet_supp}
\end{figure}

\subsection{Neural Baking}
We use a simple U-Net~\cite{UNet} as shown in Fig.~\ref{fig: unet_supp} for BakeNet, with an input of an 11-channel noise map sampled from a Gaussian distribution, each channel having a size of 512. The first convolutional layer increases the number of channels to 64, and the encoder of the U-Net processes the channels up to 1024, doubling the number of channels at each layer. The decoder then reduces the number of channels back to 64, and the final convolutional layer adjusts the output channels to 11. Skip connections are used between the encoder and decoder.

The 11 channels represent the following: 3 channels for scale, 3 channels for rotation, 3 channels for color, 1 channel for opacity, and 1 channel for offset. Specifically, we use 3 channels to represent the rotation in axis-angle form. 

Similar to GGHead~\cite{kirschstein2024gghead}, we apply a special normalization to the upsampled values from the output map corresponding to scale. We calculate the mean and maximum values of the unactivated scale for the avatar to be baked. Then, the sampled values $v$ are processed as follows:

\begin{align}
    s=s_{max}-\log \left( 1+\exp \left( -\left( v+s_{mean} \right) +s_{max} \right) \right),
\end{align}

\noindent where $s_{mean}$ and $s_{max}$ represent the mean and maximum values of the unactivated scale, respectively.

During training, we set the learning rate to \(1.0 \times 10^{-3}\) and use the Adam optimizer to optimize the U-Net.

\subsection{Head Completion}
We first render around the trained avatar for 30 frames. On average, DLib~\cite{dlib} deems 2 to 5 images valid. During PTI~\cite{pti}, we optimize the latent code for 200 iterations and fine-tune the generator parameters for 200 iterations.

\begin{figure}[!htbp]
    \centering
    \includegraphics[width=0.4\textwidth]{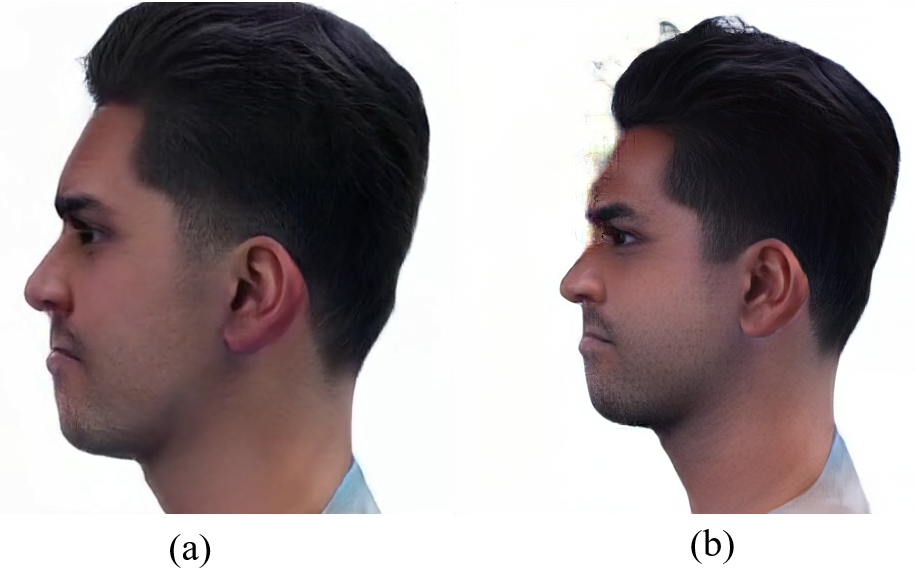}
    \caption{\textbf{Incomplete Inversion Issues}.
    In typical inversion optimization, the neck and top of the head of the portrait often fall outside the frame, as shown in (a). We obtained the result shown in (b) by adjusting the camera-to-object distance.
    }
\label{fig: pti_incomplete}
\end{figure}

We found that due to the FFHQ alignment used by SphereHead, the inversion often results in incomplete heads (see Fig.~\ref{fig: pti_incomplete}). This leads to the disappearance of some edge regions when used for completion. Since SphereHead assumes fixed camera intrinsics during training,
\begin{align}
    K=\left[ \begin{matrix}
	4.2627&		0&		0.5\\
	0&		4.2627&		0.5\\
	0&		0&		1\\
\end{matrix} \right] ,
\end{align}
directly modifying the camera intrinsics leads to poor out-of-domain results. We found a compromise by slightly increasing the camera radius from 2.7 to 3.2 while equivalently transforming the coordinates for the inverse transformation estimation to ensure the portrait appears within the viewing frustum.

After PTI is completed, we render 30 images in a full circle as pseudo-data. One potential issue is that the PTI results still differ from the real subject, and the coordinates of the monocular avatar and 3D-aware GAN are difficult to align. Therefore, we only used the latter half of the 30 images and incorporated random backgrounds during training to eliminate some artifacts.

\section{Additional Results}
\label{supp: additional_results}
\subsection{Monocular Results}
\begin{figure*}[]
    \centering
    \includegraphics[width=0.85\textwidth]{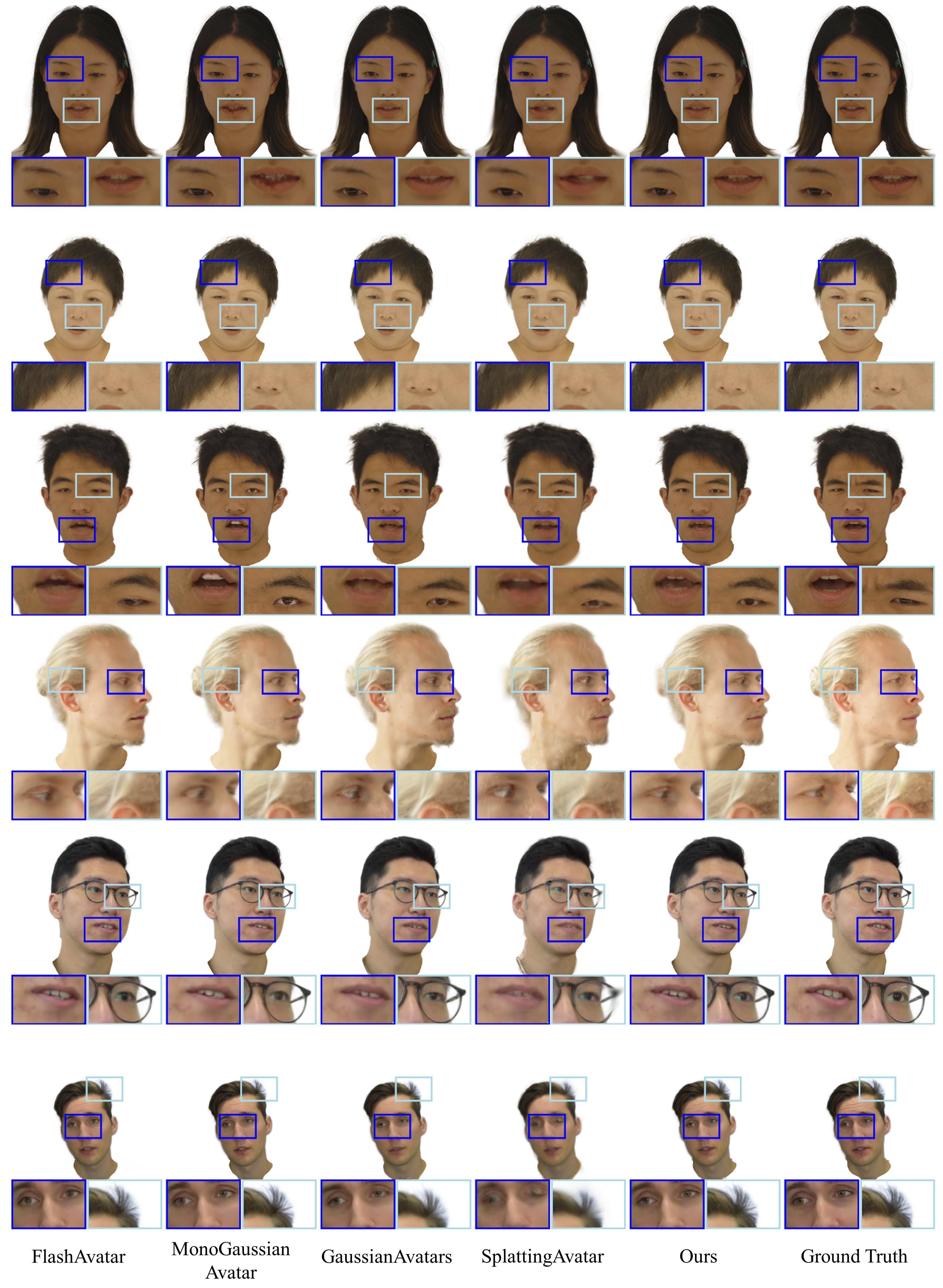}
    \caption{\textbf{More Reconstructed Results}.
    Our method excels at capturing fine structures and preserving high-frequency details (\textit{e.g.}, eyebrows, hair strands, eyeglass frames, and pupil colors.).
    }
\label{fig: mono_recon_supp}
\end{figure*}

We provide the quantitative results for each subject in Tab.~\ref{tab: insta_table} and Tab.~\ref{tab: other_table}, and more qualitative results are presented in Fig.~\ref{fig: mono_recon_supp}. Our method demonstrates superior performance across multiple datasets. 

Other methods, such as FlashAvatar, achieve excellent LPIPS scores on the INSTA dataset but perform poorly on the PointAvatar dataset, which contains complex poses and expressions. We attribute this to the deformation MLP in FlashAvatar overfitting the training set. In contrast, our method mitigates this tendency by employing a linear approach to implement personalized blendshapes, leading to better generalization.

MonoGaussianAvatar also utilizes personalized blendshapes. However, its Gaussian scales are computed through the MLP, which prefers smoothness. This smooth nature produces blurred outputs, leading to relatively high PSNR and SSIM scores but poorer LPIPS performance.

\subsection{Full-head Completion Results}
\begin{figure*}[]
    \centering
    \includegraphics[width=0.85\textwidth]{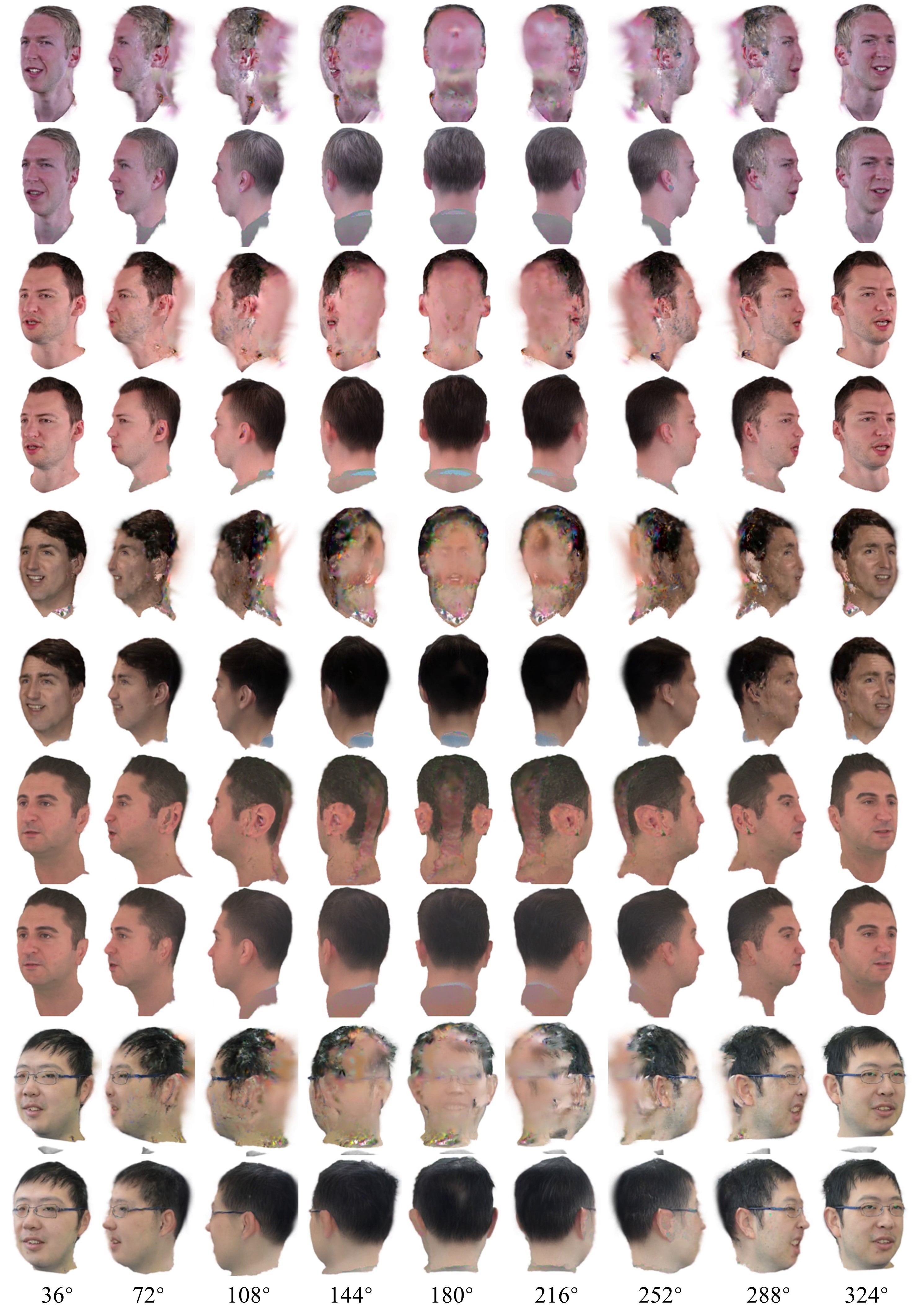}
    \caption{\textbf{More Full-head Completion Results}.
    \textit{Odd} rows display the results under novel views without applying the Full-head completion framework, while \textit{even} rows show the results after completion. Our completion framework significantly enhances rendering quality under large viewing angles.
    }
\label{fig: full_head_supp}
\end{figure*}

\begin{figure*}[]
    \centering
    \includegraphics[width=0.80\textwidth]{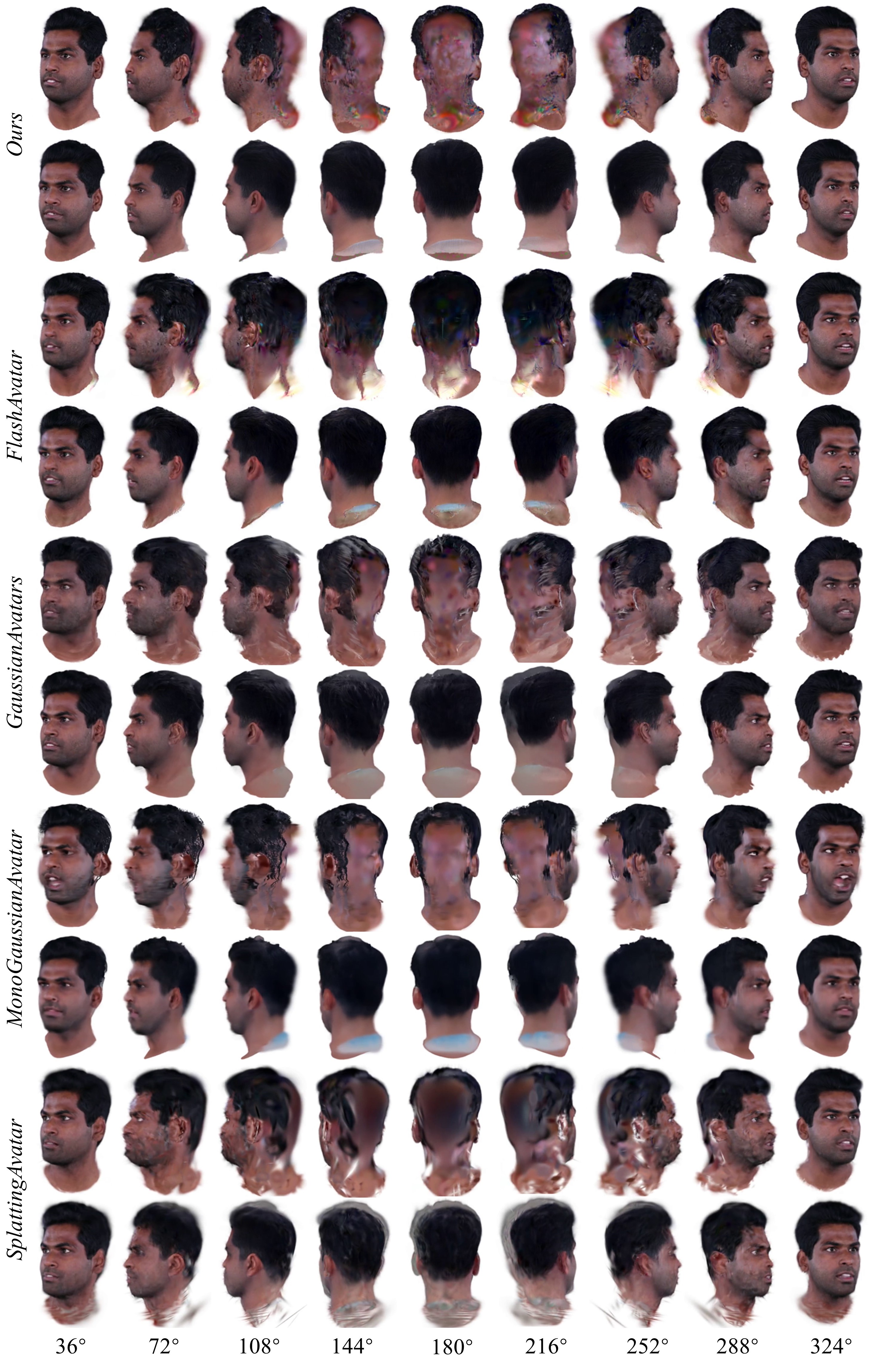}
    \caption{\textbf{Universal Completion Results}.
    \textit{Odd} rows display the results under novel views without applying the Full-head completion framework, while \textit{even} rows show the results after completion. Our completion framework applies to various monocular reconstruction methods.
    }
\label{fig: full_head_each_supp}
\end{figure*}

We provide additional results of full-head completion in Fig.~\ref{fig: full_head_supp}. Since monocular videos lack supervision for side and back views, novel views at large angles tend to perform poorly before completion. After applying the completion framework, plausible rendering results are achieved across most angles. Furthermore, we extend the completion framework to other methods. As shown in Fig.~\ref{fig: full_head_each_supp}, these methods also yield reasonable results after applying the completion framework.

We observed that for methods allowing free movement of Gaussians (\textit{e.g.}, GaussianAvatars, SplattingAvatar), misalignment artifacts are more severe. This is because the overly flexible Gaussians overfit to misaligned views. However, these methods still achieve relatively satisfactory completion results.

\subsection{Cross-reenactment Results}
\begin{figure*}[]
    \centering
    \includegraphics[width=0.98\textwidth]{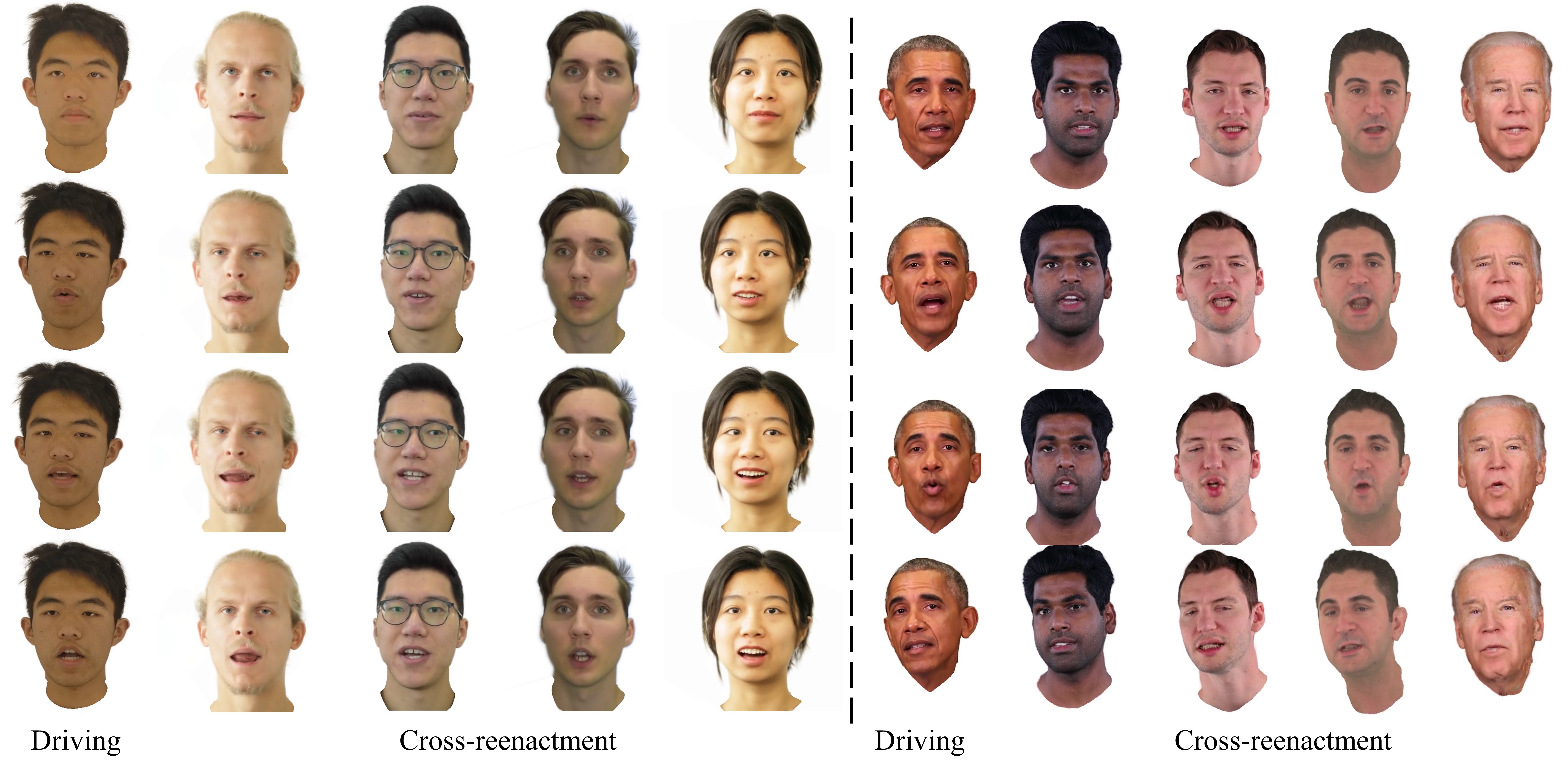}
    \caption{\textbf{Cross-reenactment Results}.
    We use the expression and pose sequences from the driving source to animate different subjects, enabling the transfer of dynamic facial expressions and poses across various avatars.
    }
\label{fig: cross_reenact_supp}
\end{figure*}

We present the results of cross-reenactment in Fig.~\ref{fig: cross_reenact_supp}. We achieve face reenactment by transferring the expression and pose of the driving avatar to different subjects. Under monocular video settings, the shape parameters and expression are not well decoupled. To achieve effective transfer, we need to compute the delta of the expression between the driving avatar and the target avatar when both exhibit a neutral expression.

\subsection{Editing Results}
\begin{figure*}[]
    \centering
    \includegraphics[width=0.92\textwidth]{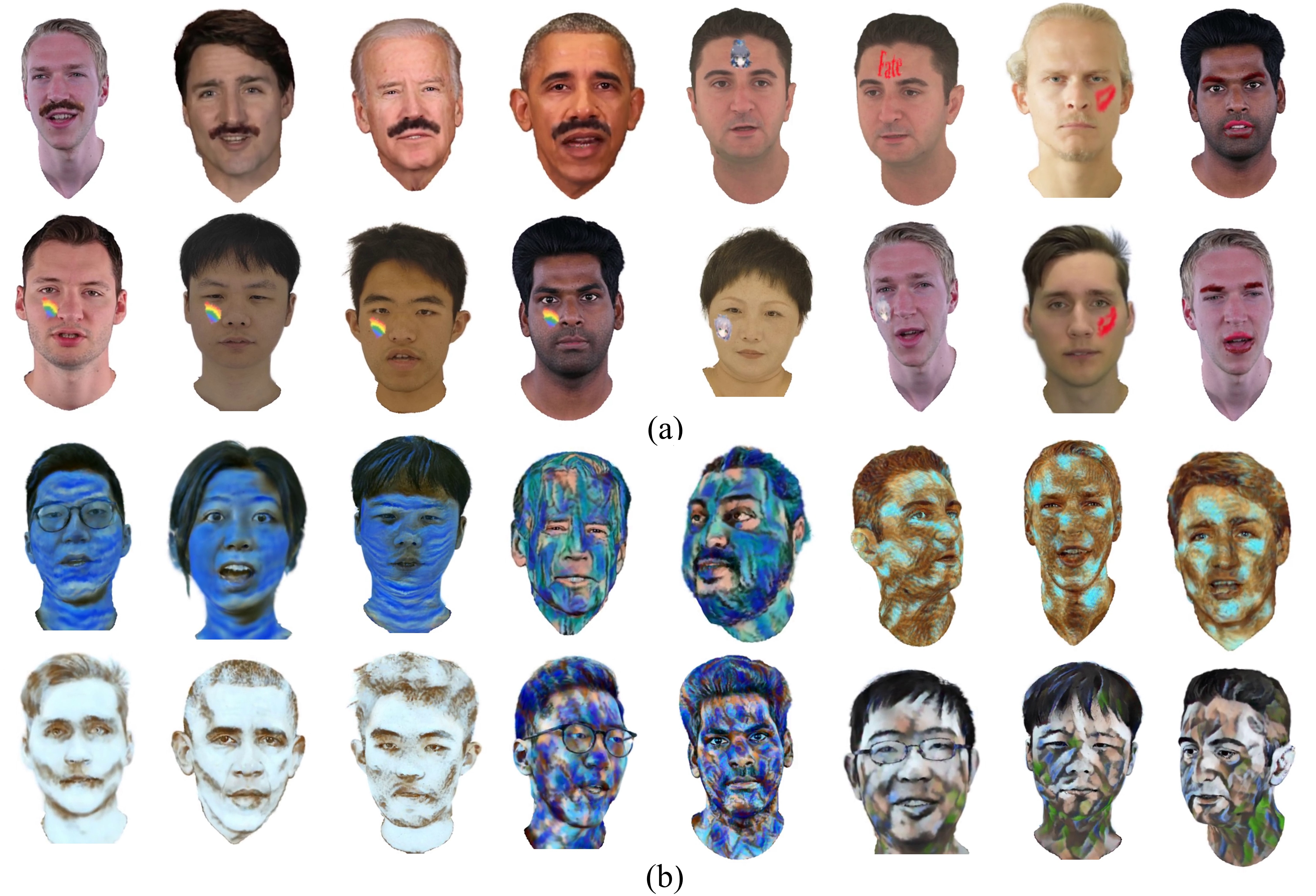}
    \caption{\textbf{Editing Results}.
    In (a), we show several results of directly editing the texture map by adding stickers, such as anime portraits, rainbows, kisses, mustaches, and logos. In (b), we present the results of applying style transfer to the texture map.
    }
\label{fig: edit_supp}
\end{figure*}

More textural editing results are shown in Fig.~\ref{fig: edit_supp}. In sticker editing, we manually craft stickers with simple patterns and corresponding masks, applying them directly to the color texture map. In style transfer, we use off-shelf and classic style transfer models~\cite{fast_neural_style_transfer} to transfer the texture map. Since we do not employ non-zero-order spherical harmonic coefficients, the results are inherently multi-view consistent after editing. Compared to methods that require pre-trained models and optimize inconsistent editing results, direct editing on the texture map is a faster and easy-to-use approach.

\begin{figure*}[!htbp]
    \centering
    \includegraphics[width=0.99\textwidth]{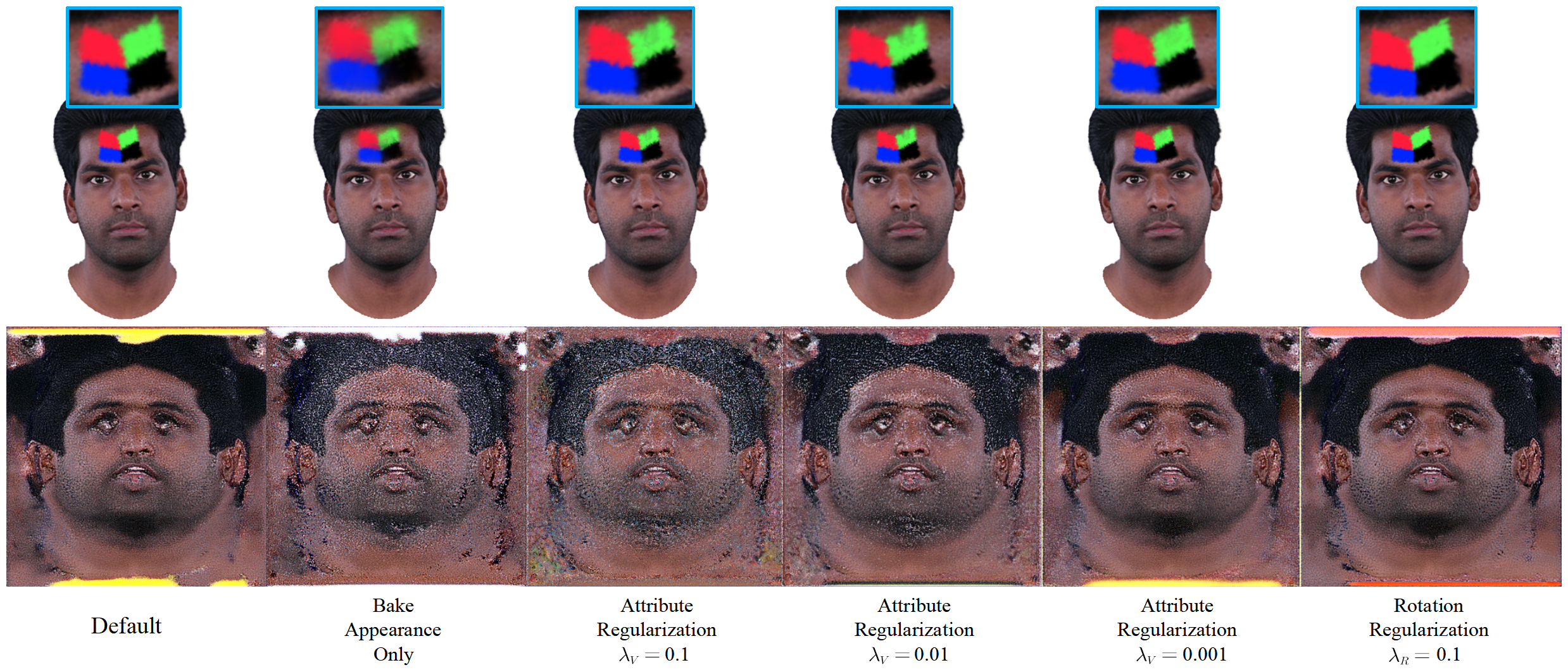}
    \caption{\textbf{Neural Baking Trade-off}.
    We visualize the color texture maps produced by neural baking under different settings and the results after editing with a checking sticker.
    }
\label{fig: trade_off_supp}
\end{figure*}

\section{Neural Baking Trade-off}
\label{supp: trade-off}

As reported in the main content and Tab.~\ref{tab: insta_table}, \ref{tab: other_table}, neural baking causes certain metric degradation compared to the avatars optimized in a point-wise manner. We found that this is because convolutional neural networks (CNN) struggle to fit the complex distribution of Gaussian geometry (scale, rotation, and offset) in the UV space. Several experiments are designed to illustrate this observation.

\noindent \textbf{Bake Appearance Only}
We only use neural baking to obtain texture maps for color and opacity, while the scale, rotation, and offset are retained from the pre-trained avatar.

\noindent \textbf{Attribute Regularization}
We minimize the difference between the attributes sampled from the BakeNet output and the corresponding attributes of the pre-trained avatar:
\begin{align}
    \mathcal{L} _V=\left\| v_{\ast}-\bar{v}_{\ast} \right\| _2,
\end{align}
where $v$ denotes sampled values and $\ast$ refers to Gaussian attributes. We add this regularization term to the baking training objective with a strength of $\lambda_V$.

\noindent \textbf{Rotation Regularization}
We impose a regularization term on the sampled rotation. Since our rotation is relative to the local triangle, we enforce the rotation around its $x$-axis and $y$-axis to be close to 0. This encourages the Gaussian rotation around the face's normal direction:
\begin{align}
    \mathcal{L} _R=\left\| r_x \right\| _2+\left\| r_y \right\| _2,
\end{align}
where $r_x$ and $r_y$ are rotations in axis-angle representation.

\begin{table}[]
\centering
\caption{The quantitative results of the neural baking trade-off in \textit{bala} case. \colorbox{blue!50} {blue} indicate the best.}
\scalebox{0.95}{\begin{tabular}{llll}
\toprule
                   & PSNR↑ & SSIM↑  & LPIPS↓ \\ \hline
Default               & \cb{29.27} & 0.9278 & 0.0584 \\
Bake App. Only        & 28.76 & \cb{0.9298} & \cb{0.0522} \\
Attribute Regu. $\lambda_{V}=0.1$ & 29.13 & 0.9239 & 0.0582 \\
Attribute Regu. $\lambda_{V}=0.01$ & 29.18 & 0.9264 & 0.0583 \\
Attribute Regu. $\lambda_{V}=0.001$ & 29.18 & 0.9268 & 0.0581 \\
Rotation  Regu. $\lambda_{R}=0.1$ & 29.22 & 0.9272 & 0.0592 \\ \bottomrule

\end{tabular}}
\label{tab: trade_off_supp}
\end{table}

Quantitative and qualitative results are shown in Tab.~\ref{tab: trade_off_supp} and Fig.~\ref{fig: trade_off_supp}. When we bake only the color and opacity while retaining the pre-trained Gaussian geometric attributes, the LPIPS metric improves. However, it leads to noisy texture maps and blurry edited stickers. A straightforward idea is to make the baked attributes approximate the pre-trained ones. We conduct experiments under three levels of $\lambda_V$, but the results show that the metrics are still decreased even at the cost of degrading the texture maps. This suggests that CNN struggles to fit the complex geometric distribution of Gaussian attributes in UV space. We believe this is because the attributes describing the Gaussian geometry lack local similarity, making them ill-suited for learning with CNN. Additionally, we introduce a rotation regularization term during baking, which worsens LPIPS but improves the quality of the texture maps and editing effects.

These experiments demonstrate that we can flexibly balance rendering quality and texture map quality in practice. If better rendering quality is desired, we can opt not to bake the geometric attributes of Gaussians. Conversely, if smoother texture maps or better editing effects are desired, applying regularization terms, such as rotation regularization, can make the Gaussians more isotropic and closer to the surface, thereby resulting in smoother texture maps.

\section{Failure Case and Limitation}
\label{supp: failure}

\begin{figure}[]
    \centering
    \includegraphics[width=0.4\textwidth]{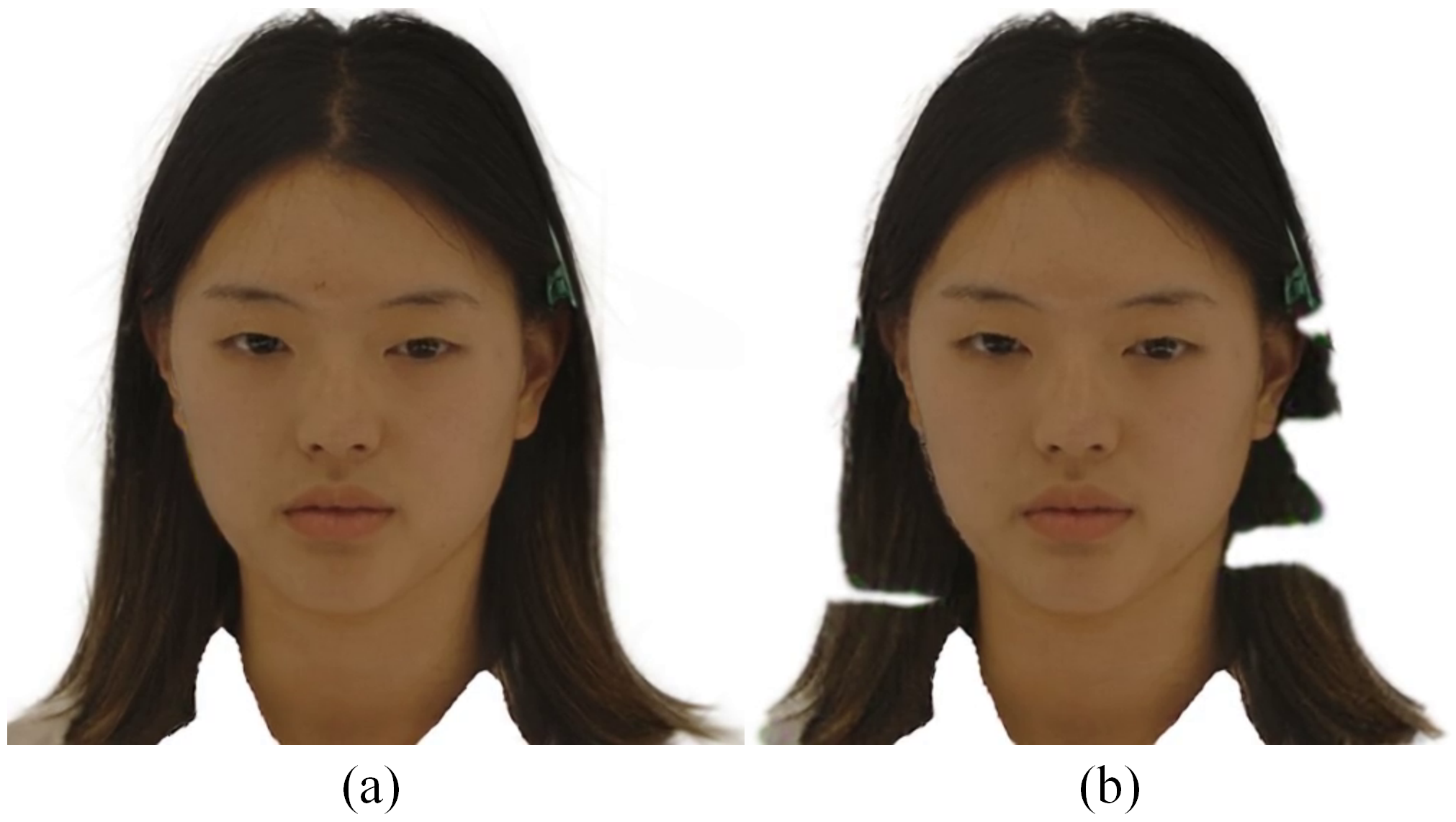}
    \caption{\textbf{Neural Baking Failure}.
    For long hair subjects, as in (a), direct neural baking will damage the fine geometry of the Gaussians composing the hair as in (b).
    }
\label{fig: fail_1}
\end{figure}

Our neural baking and full-head completion still have limitations. As mentioned in Sec.~\ref{supp: trade-off}, since CNN is tricky to construct Gaussian geometry, neural baking may fail for intricate geometry. For instance, in the case of the woman with long hair shown in Fig.~\ref{fig: fail_1}, the hair requires Gaussians with delicate scale and rotation. However, neural baking makes it difficult to recover the desired geometry.

\begin{figure}[]
    \centering
    \includegraphics[width=0.4\textwidth]{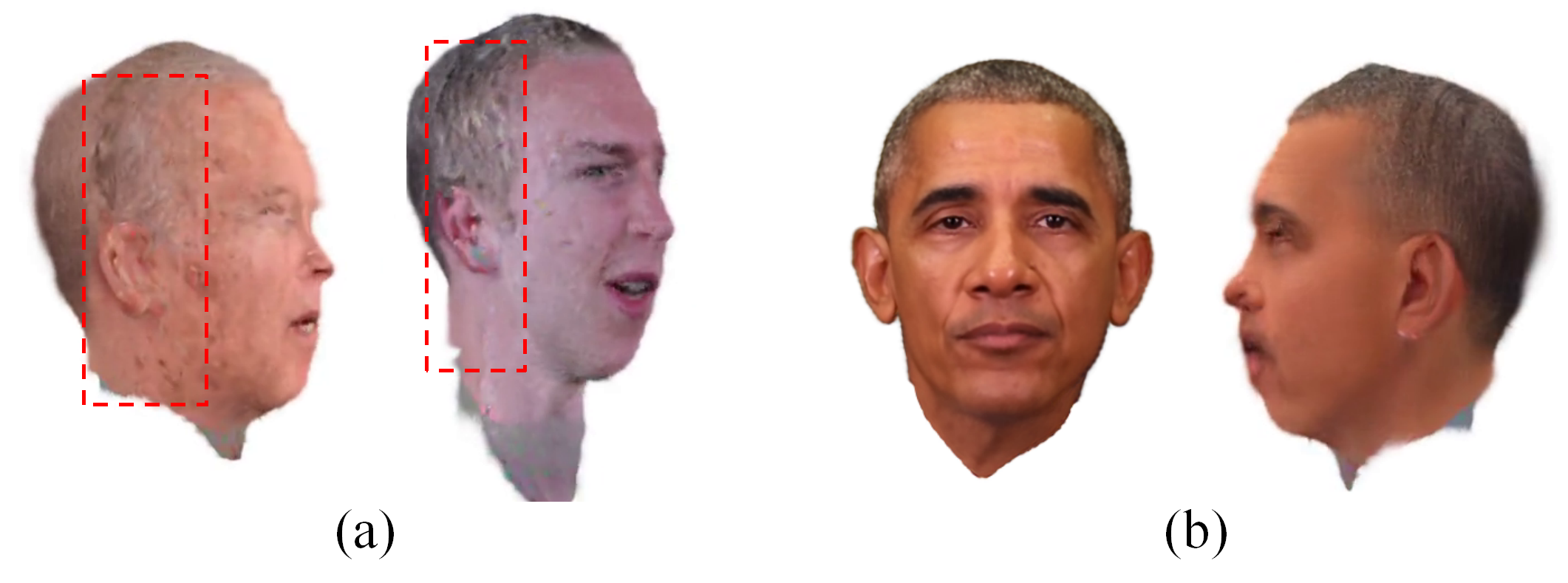}
    \caption{\textbf{Full-head Completion Failure}.
    Since the PTI results still differ from the real avatar, artifacts appear at the junction, as shown in the red box in (a). And for avatars with almost no side view in the training data, as shown in (b), it is difficult to estimate the exact geometry during PTI, leading to the identity change in the side view.
    }
\label{fig: fail_2}
\end{figure}

In full-head completion, we are training the unobserved view with pseudo images and the frontal view with real images. Artifacts, as shown in Fig.~\ref{fig: fail_2} (a), may appear in a certain side-view angle due to the transition between the two regions. Additionally, for subjects that have almost no side view in monocular videos (\textit{e.g.}, Internet video focusing on talking), PTI does not estimate the head with the correct geometry, resulting in identity change.

\section{Noisy Pose Simulation}
\label{supp: noise_pose}

\begin{figure*}[]
    \centering
    \includegraphics[width=0.8\textwidth]{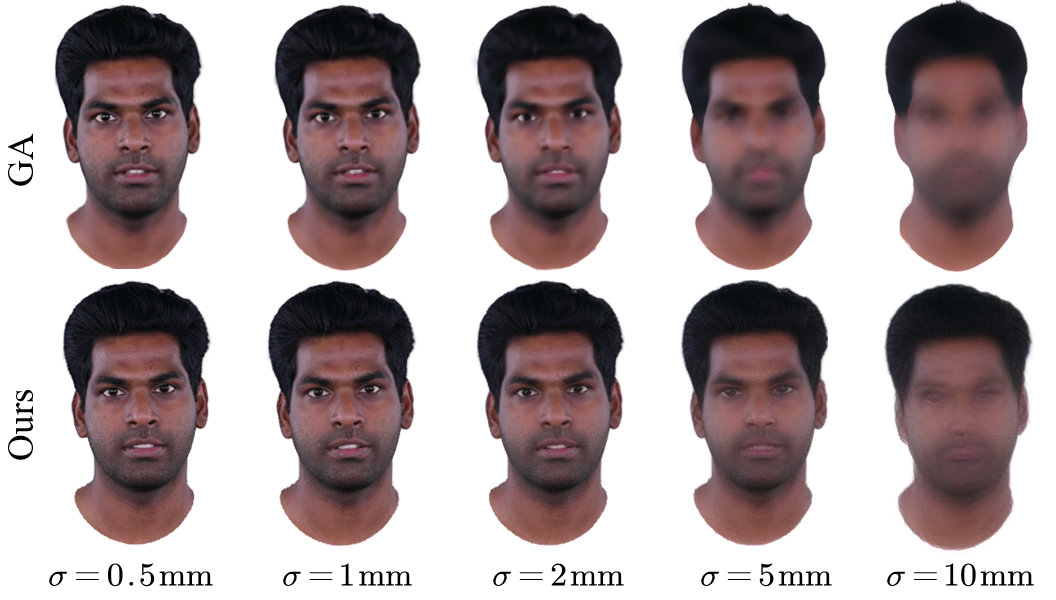}
    \captionsetup{justification=centering}
    \caption{\textbf{Robustness to Imperfect Poses} We add noise to camera translation to simulate less well-processed datasets. Note that 1 mm in the figure approximately corresponds to 1 cm in the real world.}
    \label{fig: noisy}
\end{figure*}

To train head avatars from monocular videos, we require frame-by-frame RGB images along with the corresponding tracked coefficients. We further evaluate the differences between our method and GaussianAvatars when the camera translation is imperfect. We add Gaussian noise with varying $\sigma$ to camera translations to simulate real-captured data with inaccurate tracking. Fig.~\ref{fig: noisy} shows our method is more robust than GA to such conditions. We attribute this to the regularization of the UV embedding, which constrains the Gaussians from freely moving to a blurred average solution.

\section{Computational Efficiency}
\label{supp: compute}


In Tab.~\ref{tab: compute}, we supplement the training time and rendering FPS under identical hardware conditions.
Our method outperforms other UV space-based methods (FA, SA) regarding shorter training time and higher FPS. Compared to GA, our method achieves comparable efficiency with superior rendering quality.

We also measure the average running time of each part in our proposed method on the INSTA dataset. We just fine-tune for $1$ epoch during completion and $5$ epochs during baking, with training times ranging from 0.5 to 1 hour, depending on whether only the frontal face or the entire head is baked. The average running time is shown in Tab.~\ref{tab: more_time_tab}

\begin{table}[]
\captionsetup{justification=centering}
\caption{Running Time on Optional Parts.}
\scalebox{0.8}{
\begin{tabular}{ccccc}
\hline
     & Mono. Recon. & Pseudo Gen.  & Completion   & Neural Baking \\ \hline
Time & $\sim$ 1.0h  & $\sim$ 10min & $\sim$ 15min & $\sim$ 0.5h/1.0h  \\ \hline
\end{tabular}
}
\label{tab: more_time_tab}
\end{table}

\begin{table*}[]
\captionsetup{justification=centering}
\caption{Evaluation on Computational Efficiency.}
\vspace{-0.1in}
\centering
\scalebox{1.0}{
\begin{tabular}{c|cccccccccccc}
\toprule
\multirow{2}{*}{Datasets} & \multicolumn{3}{c|}{INSTA Dataset}                        & \multicolumn{3}{c|}{IMAvatar Dataset}                     & \multicolumn{3}{c|}{NerFace Dataset}                      & \multicolumn{3}{c}{EmoTalk3D Dataset}        \\ \cline{2-13} 
                          & GS num. & {\scriptsize \begin{tabular}[c]{@{}c@{}}Training\\  time\end{tabular}} & \multicolumn{1}{l|}{FPS} & GS num. & {\scriptsize \begin{tabular}[c]{@{}c@{}}Training\\  time\end{tabular}} & \multicolumn{1}{l|}{FPS} & GS num. & {\scriptsize \begin{tabular}[c]{@{}c@{}}Training\\  time\end{tabular}} & \multicolumn{1}{l|}{FPS} & GS num. & {\scriptsize \begin{tabular}[c]{@{}c@{}}Training\\  time\end{tabular}} & FPS \\ \hline
FA                        & 16k         & 1.4h      & 190               & 16k                          & 4.5h         & 208      & 16k               & 5.5h                          & 206         & 16k      & 0.5h               & 214                               \\
SA                        & 558k±188k         & 1.2h      & 106               & 617k±274k                          & 8.0h         & 109      & 497k±142k               & 9.6h                          & 112         & 489k±171k      & 1.0h               & 116                             \\
MGA                       & 100k         & 7.2h      & 16               & 100k                          & 13h         & 16      & 100k               & 14h                          & 16         & 100k      & 7.5h               & 16                              \\
GA                        & 72k±33k         & 0.4h      & 212               & 38±14k                          & 3.1h         & 228      & 31k±7k               & 4.2h                          & 223         & 55±12k      & 0.8h               & 229                              \\
Ours                      & 49k±6k         & 1.0h      & 203               & 38k±6k                          & 3.7h         & 216      & 42k±0.5k               & 4.5h                          & 215         & 58k±2k      & 0.5h               & 216                              \\ \bottomrule
\end{tabular}
}
\label{tab: compute}
\end{table*}

\section{More Ablations}
\label{supp: more_abl}

As shown in Tab.~\ref{tab: more_abl}, we introduce more ablation settings on two representative datasets and further report the number of Gaussian in different settings. We additionally conduct experiments as \textit{w/o densify$^*$}, where $^*$ indicates that Gaussians are removed based on opacity criteria. The suboptimal results further highlight the effectiveness of sampling-based densification. Moreover, we align our method with densification strategies based on SA and GA. GA-based densification tends to produce blurrier results, while SA-based densification introduces too many redundant Gaussians.

And we supplement more experiments comparing \textit{Two-stage} and \textit{One-stage}. Concretely, we find baking only the appearance (denoted as \textit{Two-stage baking App.}) improves rendering quality compared to \textit{Two-stage baking} but causes blurred editing effects, which is visualized and discussed in Sec.~4 of our supplementary material. Besides, we additionally report GS numbers in Tab.~\ref{tab: more_abl} to show that \textit{Two-stage baking} leveraging the evolved distribution achieves comparable performance with much fewer Gaussians than \textit{One-stage baking} that uses initialized uniform distribution.

\begin{table*}[!htbp]
\captionsetup{justification=centering}
\caption{Ablation Study in \textit{yufeng} and \textit{bala}.}
\vspace{-0.1in}
\centering
\scalebox{1.0}{
\begin{tabular}{ccccccccc}
\hline
                     & \multicolumn{4}{c}{yufeng}                                                                             & \multicolumn{4}{c}{bala}                                                                             \\ \cline{2-9} 
                     & PSNR↑ & SSIM↑  & LPIPS↓ & GS num.                                                                      & PSNR↑ & SSIM↑ & LPIPS↓ & GS num.                                                                     \\ \hline
Ours                 & 29.36 & 0.9239 & 0.0694 & 30k                                                                             & 29.23      & 0.9329      & 0.0507       & 54k                                                                            \\
w/o densify$^*$          & 28.81  & 0.9195 & 0.0799 & 14k                                                                            & 28.83      & 0.9309      & 0.0526       & 45k                                                                            \\
w/o densify          & 29.13 & 0.9217 & 0.0740 & 65k                                                                             & 28.91      & 0.9311      & 0.0528       & 65k                                                                            \\
w/o $\Delta \mathcal{E}$ and $\Delta \mathcal{P}$ & 24.78 & 0.8820 & 0.1112 & 33k                                                                             & 24.46      & 0.9015      & 0.1081       & 54k                                                                            \\
w/ GA densify        & 29.62      & 0.9327       & 0.0941       & 80k                                                                             & 26.89      & 0.9270      & 0.0966       & 118k                                                                            \\
w/ SA densify        & 27.74      & 0.8699       & 0.1896       & 803k                                                                             & 26.37      & 0.8417      & 0.1827       & 917k                                                                            \\ \hline
Two-stage baking     & 27.78 & 0.9104 & 0.0979 & 30k                                                                             & 29.27      & 0.9278      & 0.0584       & 54k                                                                            \\
Two-stage baking App.     & 28.84  & 0.9190 & 0.0797 & 30k                                                                                   & 29.53      & 0.9298      & 0.0522       & 54k                                                                            \\
One-stage baking     & 27.42 & 0.9085 & 0.1088 & 65k                                                                             & 29.12      & 0.9208      & 0.0602       &  65k                                                                           \\
Decode only          & 25.56 & 0.8878 & 0.1506 & 30k                                                                             & 28.25      & 0.9071      & 0.0827       & 54k                                                                            \\ \hline
\end{tabular}
}
\label{tab: more_abl}
\end{table*}

\section{Ethics}
\label{supp: ethics}
We used four subjects from EmoTalk3D~\cite{he2024emotalk3d}, with all participants signing the consent for using their videos in this research and publication. Data from consenting subjects will be made publicly available. Our method generates realistic and animatable head avatars, enabling the creation of videos of real people performing synthetic poses and expressions. We strictly oppose any misuse of this work to create deceptive content intended to spread misinformation or damage reputations.

\begin{table*}[]
\centering
\caption{Full comparison of quantitative results with state-of-the-art methods on INSTA dataset. \colorbox{blue!50} {blue} and \colorbox{blue!20}{lightblue} indicate the 1st and 2nd best.}
\label{tab: insta_table}
\centering
\scalebox{0.92}{
\begin{tabular}{ll|llllllllll}
\toprule
\multicolumn{2}{c|}{\multirow{2}{*}{Datasets}}                   & \multicolumn{10}{c}{INSTA Dataset}                                                                                                                                                                                                                                                                              \\ \cline{3-12} 
\multicolumn{2}{c|}{}                                            & \multicolumn{1}{c}{bala}   & \multicolumn{1}{c}{biden}  & \multicolumn{1}{c}{justin}  & \multicolumn{1}{c}{malte\_1} & \multicolumn{1}{c}{marcel}  & \multicolumn{1}{c}{nf\_01}   & \multicolumn{1}{c}{nf\_03}   & \multicolumn{1}{c}{obama}    & \multicolumn{1}{c}{person4} & \multicolumn{1}{c}{wojtek\_1} \\ \midrule
\multicolumn{1}{l|}{\multirow{5}{*}{PSNR↑}} & FlashAvatar        & 28.04                           & 29.63                           & \cb{27.42}                            & 27.54                             & \cb{29.32}                            & 24.47                                   & \clb{27.24}                             & 26.92                             & 21.65                              & 30.66                              \\
\multicolumn{1}{l|}{}                       & SplattingAvatar    & 27.58                           & 28.64                           & 26.43                            & \clb{28.01}                             & 28.05                            & 24.11                                   & 26.02                             & 25.46                             & 20.84                              & 31.23                              \\
\multicolumn{1}{l|}{}                       & MonoGaussianAvatar & \cb{28.95}                          & \clb{29.97}                          & \clb{27.28}                            & 27.93                             &  28.84                           & \clb{24.58}                                    & 27.22                             & \cb{27.39}                             & \clb{21.98}                             & 29.90                              \\
\multicolumn{1}{l|}{}                       & GaussianAvatars    & 27.99                           & 28.52                           & 26.70                            & 27.53                             & \clb{29.10}                            & 23.80                                   & 26.00                             & 25.12                             & 20.58                             & \cb{31.27}                              \\
\multicolumn{1}{l|}{}                       & Ours               & \clb{28.17}                           & \cb{30.06}                           & 27.11                            & \cb{28.40}                             & 29.07                           & \cb{24.69}                                    & \cb{27.46}                             & \clb{27.04}                             & \cb{22.01}                              & \clb{31.25}                              \\ \hline
\multicolumn{1}{l|}{}                       & Ours (Baked)               & 29.42                           & 30.37                           & 27.75                            & 28.34                             & 28.38                           & 24.91                                    & 27.43                             & 28.01                             & 21.92                              & 31.49                              \\ \hline
\multicolumn{1}{l|}{\multirow{5}{*}{SSIM↑}} & FlashAvatar        & 0.9170                           & 0.9595                           & 0.9580                            & 0.9310                             & 0.9328                            & 0.9303                             & 0.9227                             & 0.9413                             & 0.9009                              & 0.9530                              \\
\multicolumn{1}{l|}{}                       & SplattingAvatar    & 0.9168                           & 0.9501                           & 0.9553                            & 0.9352                             & 0.9265                            & 0.9201                             & 0.9137                             & 0.9341                             & 0.8959                              & 0.9563                              \\
\multicolumn{1}{l|}{}                       & MonoGaussianAvatar & \clb{0.9297}                          &  \clb{0.9599}                          & 0.9601                            &  0.9340                            & 0.9316                           & \clb{0.9320}                               & 0.9200                             & \clb{0.9482}                             & \clb{0.9090}                              & 0.9480                              \\
\multicolumn{1}{l|}{}                       & GaussianAvatars    & \cb{0.9397}                           & 0.9548                           & \cb{0.9609}                            & \clb{0.9401}                             & \cb{0.9394}                            & 0.9286                             & \cb{0.9257}                             & 0.9398                             & 0.9032                              & \cb{0.9635}                              \\
\multicolumn{1}{l|}{}                       & Ours               & 0.9267                           & \cb{0.9659}                           & \clb{0.9606}                            & \cb{0.9434}                             & \clb{0.9371}                            & \cb{0.9344}                             & \clb{0.9253}                             & \cb{0.9524}                             & \cb{0.9097}                              & \clb{0.9600}                              \\ \hline
\multicolumn{1}{l|}{}                       & Ours (Baked)               & 0.9285                           & 0.9672                           & 0.9638                            & 0.9424                             & 0.9356                           & 0.9347                                    & 0.9231                             & 0.9547                             & 0.9086                              & 0.9602                              \\ \hline
\multicolumn{1}{l|}{\multirow{5}{*}{LPIPS↓}} & FlashAvatar        & \cb{0.0484}                           & \cb{0.0267}                           & \cb{0.0410}                            & \cb{0.0418}                             & \cb{0.0757}                            &  \cb{0.0768}                            & \cb{0.0601}                             & \cb{0.0375}                             & \clb{0.1403}                              & \cb{0.0296}                              \\
\multicolumn{1}{l|}{}                       & SplattingAvatar    & 0.1284                           &  0.0694                          &  0.0902                           & 0.0838                             & 0.1330                            & 0.1385                             & 0.1286                             & 0.0820                             & 0.1992                              & 0.0659                              \\
\multicolumn{1}{l|}{}                       & MonoGaussianAvatar & 0.0969                           &  0.0547                          & 0.0585                            & 0.0758                             & 0.1352                            & 0.1146                             & 0.0980                             & 0.0532                             & 0.1471                              & 0.0534                              \\
\multicolumn{1}{l|}{}                       & GaussianAvatars    & 0.0618                           & 0.0458                           & 0.0616                            & 0.0607                             & 0.0943                            & 0.1025                             & 0.0827                             & 0.0606                             & 0.1626                              & 0.0445                              \\
\multicolumn{1}{l|}{}                       & Ours               & \clb{0.0534}                           & \clb{0.0341}                           & \clb{0.0445}                            & \clb{0.0414}                             & \clb{0.0760}                            & \clb{0.0811}                             & \clb{0.0674}                             & \clb{0.0426}                             & \cb{0.1298}                              & \clb{0.0324}                              \\ \hline
\multicolumn{1}{l|}{}                       & Ours (Baked)               & 0.0583                           & 0.0371                           & 0.0448                            & 0.0451                             & 0.0841                           & 0.0859                                    & 0.0755                             & 0.0422                             & 0.1312                              & 0.0346                              \\ \bottomrule

\end{tabular}
}


\end{table*}

\begin{table*}[!htbp]
\centering
\caption{Full comparison of quantitative results with state-of-the-art methods on the PointAvatar dataset, NerFace dataset, and Emotalk3D dataset. \colorbox{blue!50} {blue} and \colorbox{blue!20}{lightblue} indicate the 1st and 2nd best.}
\label{tab: other_table}
\scalebox{0.85}{
\begin{tabular}{ll|llllllllll}
\toprule
\multicolumn{2}{c|}{\multirow{2}{*}{Datasets}}                   & \multicolumn{3}{c|}{PointAvatar Dataset}                                                                  & \multicolumn{3}{c|}{NerFace Dataset}                                                                         & \multicolumn{4}{c}{Emotalk3D Dataset}                                                                                             \\ \cline{3-12} 
\multicolumn{2}{c|}{}                                            & \multicolumn{1}{c}{yufeng} & \multicolumn{1}{c}{marcel} & \multicolumn{1}{c}{soubhik} & \multicolumn{1}{c}{person1}  & \multicolumn{1}{c}{person2} & \multicolumn{1}{c|}{person3} & \multicolumn{1}{c}{subject1} & \multicolumn{1}{c}{subject2} & \multicolumn{1}{c}{subject3}  & \multicolumn{1}{c}{subject4}  \\ \midrule
\multicolumn{1}{l|}{\multirow{5}{*}{PSNR↑}} & FlashAvatar        & 25.85                           & 26.67                           & 26.84                            & 30.44                             & 31.47                            & 32.24                                   & 25.28                             & 29.96                             & 21.41                              & \clb{25.28}                              \\
\multicolumn{1}{l|}{}                       & SplattingAvatar    & 25.09                           & 25.57                           & 23.61                            & 27.35                             & 28.58                            & 32.10                                   & 25.16                              & 28.23                             & 20.98                              & 23.81                              \\
\multicolumn{1}{l|}{}                       & MonoGaussianAvatar & \clb{28.50}                           & \clb{27.01}                           & \clb{28.98}                            & \clb{32.09}                            & \cb{33.89}                            & \cb{35.35}                                    & \clb{25.31}                             & \clb{30.22}                             & \clb{21.46}                              & 24.74                              \\
\multicolumn{1}{l|}{}                       & GaussianAvatars    & 25.12                           & 25.15                          & 23.27                            & 27.03                             & 28.62                            & 31.54                                    & 25.12                             & 27.71                             & 20.77                              & 23.10                              \\
\multicolumn{1}{l|}{}                       & Ours               & \cb{29.36}                           & \cb{27.58}                          & \cb{29.28}                            & \cb{32.91}                             & \clb{33.65}                            & \clb{34.54}                                    & \cb{26.13}                             & \cb{30.96}                             & \cb{21.56}                              & \cb{26.36}                              \\ \hline
\multicolumn{1}{l|}{}                       & Ours (Baked)               & 27.78                           & 26.37                           & 28.21                            & 31.99                             & 33.34                           & 32.45                                    & -                             & 29.70                             & 21.73                              & 26.96                              \\ \hline
\multicolumn{1}{l|}{\multirow{5}{*}{SSIM↑}} & FlashAvatar        & 0.8863                           & 0.9224                           & 0.9221                            & 0.9620                             & 0.9734                            & 0.9569                             & 0.9218                             & 0.9411                             & 0.9087                              & \clb{0.9037}                              \\
\multicolumn{1}{l|}{}                       & SplattingAvatar    & 0.8761                           & 0.9056                           & 0.8903                            & 0.9416                             & 0.9516                            & 0.9509                             & 0.9250                             & 0.9350                             & 0.9164                              & 0.9020                              \\
\multicolumn{1}{l|}{}                       & MonoGaussianAvatar & \cb{0.9259}                           & \cb{0.9374}                           &\cb{0.9448}                            & \cb{0.9719}                             & \cb{0.9812}                            & \cb{0.9763}                             & \cb{0.9434}                             & \clb{0.9453}                             & 0.9126                              & 0.8827                              \\
\multicolumn{1}{l|}{}                       & GaussianAvatars    & 0.8938                           & 0.9200                           & 0.9095                            & 0.9504                             & 0.9613                            & 0.9560                             & 0.9394                             & 0.9381                             & \clb{0.9200}                              & 0.9028                              \\
\multicolumn{1}{l|}{}                       & Ours               & \clb{0.9239}                           & \clb{0.9341}                           & \clb{0.9418}                            & \clb{0.9716}                             & \clb{0.9802}                            & \clb{0.9691}                             & \clb{0.9429}                             & \cb{0.9530}                             & \cb{0.9208}                              & \cb{0.9265}                              \\ \hline
\multicolumn{1}{l|}{}                       & Ours (Baked)               & 0.9104                           & 0.9282                           & 0.9330                            & 0.9655                             & 0.9761                           & 0.9579                                    & -                             & 0.9505                             & 0.9199                              & 0.9274                              \\ \hline
\multicolumn{1}{l|}{\multirow{5}{*}{LPIPS↓}} & FlashAvatar        & 0.1043                           & \clb{0.1021}                           & \clb{0.0607}                            & \clb{0.0377}                             & \clb{0.0217}                            & 0.0317                             & \cb{0.0668}                             & \clb{0.0435}                             & \clb{0.0827}                              & \cb{0.0787}                              \\
\multicolumn{1}{l|}{}                       & SplattingAvatar    & 0.1502                           & 0.1510                            & 0.1490                            & 0.0835                             & 0.0660                            & 0.0640                             & 0.1403                             & 0.0877                             & 0.1109                              & 0.1484                              \\
\multicolumn{1}{l|}{}                       & MonoGaussianAvatar & \clb{0.0993}                           & 0.1280                            & 0.0656                            & 0.0473                             & 0.0224                            & \cb{0.0249}                             & 0.0758                             & 0.0546                             & \cb{0.0770}                              & 0.0921                              \\
\multicolumn{1}{l|}{}                       & GaussianAvatars    & 0.1287                           & 0.1321                           & 0.1163                            & 0.0681                             & 0.0415                            & 0.0432                             & 0.0806                             & 0.0590                             & 0.0912                              & 0.1035                              \\
\multicolumn{1}{l|}{}                       & Ours               & \cb{0.0694}                           & \cb{0.0876}                           & \cb{0.0586}                            & \cb{0.0329}                             & \cb{0.0186}                            & \clb{0.0256}                              & \clb{0.0705}                             & \cb{0.0414}                             & 0.0843                              & \clb{0.0842}                              \\ \hline
\multicolumn{1}{l|}{}                       & Ours (Baked)               & 0.0979                           & 0.1142                           & 0.0740                            & 0.0460                             & 0.0240                           & 0.0418                                    & -                             & 0.0593                             & 0.0923                              & 0.0952                              \\ \bottomrule
\end{tabular}
}
\end{table*}

\end{document}